\documentclass[10pt,journal,compsoc]{IEEEtran}
\ifCLASSOPTIONcompsoc
  \usepackage[nocompress]{cite}
\else
  \usepackage{cite}
\fi
\ifCLASSINFOpdf
\else
\fi
\usepackage{parskip}
\usepackage{amssymb}
\usepackage{framed,multirow}
\usepackage{wrapfig}
\usepackage{caption}
\usepackage{dsfont}
\usepackage{latexsym}
\usepackage{array}
\usepackage{url}
\usepackage{amsmath}
\usepackage{amsthm}
\usepackage{xcolor}
\definecolor{newcolor}{rgb}{.8,.349,.1}

\usepackage{hyperref}

\usepackage[switch,pagewise]{lineno} 
\usepackage{soul}
\usepackage{proofread}
\newtheorem{theorem}{Theorem}
\newtheorem{corollary}{Corollary}

\graphicspath{{figs/}}
\newcommand{\jj}[1]{{\color{black} #1}}
\newcommand{\jz}[1]{{\color{black} #1}}

\newcommand{\re}[1]{{\color{black} #1}}

\newcommand{\fig}{Fig.}
\newcommand{\tab}{Tab.}
\newcommand{\eq}{Eq.}
\newcommand{\etal}{\emph{et al.}}
\newcommand{\ie}{\emph{i.e.}}

\usepackage{booktabs}
\usepackage{overpic}

\begin{document}
%
\title{PointNorm-Net: Self-Supervised Normal Prediction of 3D Point Clouds via Multi-Modal Distribution Estimation}

%
%
%
%

\author{Jie~Zhang,~Minghui~Nie,Changqing~Zou,~\IEEEmembership{Member,~IEEE},~Jian~Liu,  ~Ligang~Liu,~\IEEEmembership{Member,~IEEE}, and ~Junjie~Cao$^{\ast}$ \thanks{*Corresponding author},~\IEEEmembership{Member,~IEEE} 

\IEEEcompsocitemizethanks{\IEEEcompsocthanksitem Jie Zhang is with School of Mathematics, Liaoning Normal University, Dalian 116024, Liaoning, China.\protect\\
E-mail: jzhang@lnnu.edu.cn.

\IEEEcompsocthanksitem Minghui Nie is with School of Artificial Intelligence and Computer Science, Jiangnan University, Wuxi 214122, Jiangsu, China.\protect\\
E-mail: mhnie99@gmail.com

\IEEEcompsocthanksitem Changqing Zou is with State Key Lab of CAD$\&$CG at Zhejiang University and Zhejiang Lab, Zhejiang University, Hangzhou 310058, Zhejiang, China.\protect\\
E-mail: aaronzou1125@gmail.com

\IEEEcompsocthanksitem Jian Liu is with School of Software, Shenyang University of Technology, Shenyang 110870, Liaoning, China.\protect\\
E-mail: jianliu@sut.edu.cn

\IEEEcompsocthanksitem Ligang Liu is with School of Mathematical Sciences, University of Science and Technology of China,  Hefei 230026, Anhui, China.
\protect\\E-mail: lgliu@ustc.edu.cn.

\IEEEcompsocthanksitem Junjie Cao is with School of Mathematical Sciences, Dalian University of Technology, Dalian 116029, Liaoning, China.\protect\\
E-mail: jjcao1231@gmail.com.
}
}

\IEEEtitleabstractindextext
{%
\begin{abstract} 
Although supervised deep normal estimators have recently shown impressive results on synthetic benchmarks, their performance deteriorates significantly in real-world scenarios due to the domain gap between synthetic and real data. 
Building high-quality real training data to boost those supervised methods is not trivial because point-wise annotation of normals for varying-scale real-world 3D scenes is a tedious and expensive task. This paper introduces PointNorm-Net, the first self-supervised deep learning framework to tackle this challenge. The key novelty of PointNorm-Net is a three-stage multi-modal normal distribution estimation paradigm that can be integrated into either deep or traditional optimization-based normal estimation frameworks. 
Extensive experiments show that our method achieves superior generalization and outperforms state-of-the-art conventional and deep learning approaches across three real-world datasets that exhibit distinct characteristics compared to the synthetic training data.

\end{abstract}

}

\maketitle
\IEEEraisesectionheading{\section{Introduction}}
\IEEEPARstart{P}{oint} clouds have emerged as a fundamental 3D representation in a wide range of applications including reverse engineering, indoor scene modeling, robot grasping, and autonomous driving. The surface normal of point clouds, which defines the first-order local structure of the underlying surface, plays a crucial role in these applications\cite{Rui2023GCNO, MultiNormal2019, HF-cubes, PCPNet:2018}. A large number of deep neural networks have been proposed for accurate normal estimation \cite{zhou2022refine, DeepFit, Adafit, NestiNet2019, Geometry-Guided, PCPNet:2018, DeepIterative, lihsurf, MDRNet}, and these networks outperform conventional methods \cite{MultiNormal2019, HF-cubes, least_squares, Zhang13, Bao, Jets, HoppeDDMS92} on synthesized benchmark datasets. However, their performance on raw point clouds drops significantly because they are not trained with real data and there exists a domain gap between real and synthetic data. 

\begin{figure*}[h]
\centering
\vspace{-0.1cm} 
\includegraphics[width=1\textwidth, keepaspectratio]{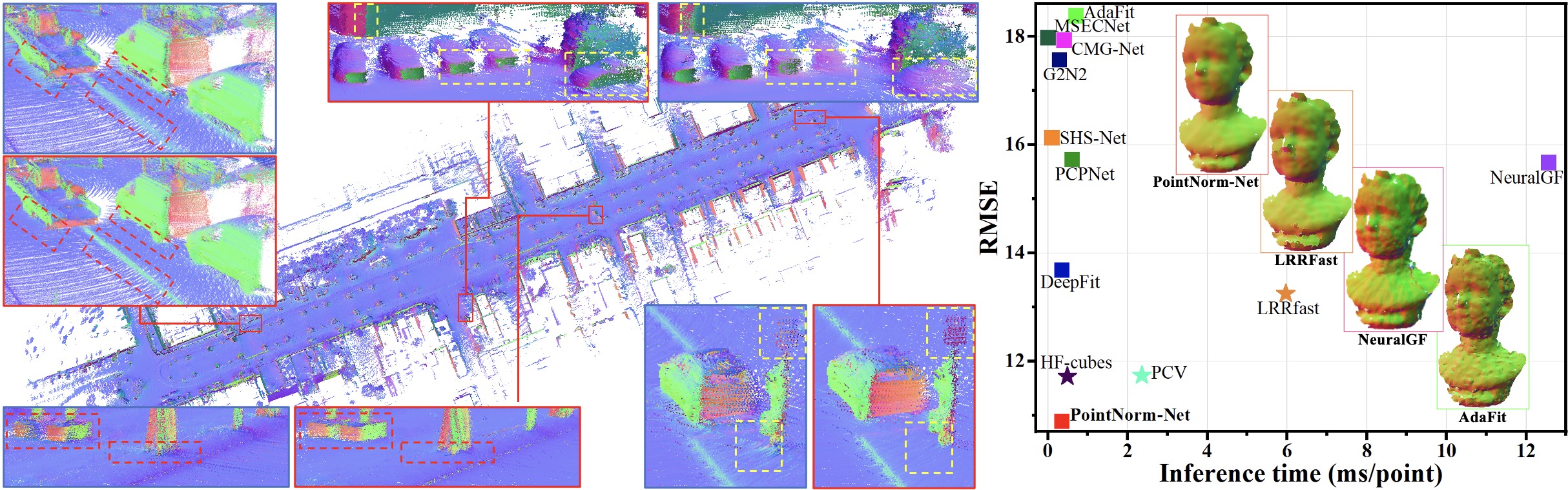}
\vspace{-0.5cm} 
\caption{PointNorm-Net demonstrates better performance in comparison to traditional optimization-based methods and supervised deep normal estimators when dealing with real-world point cloud datasets, such as LiDAR sequence 06 of KITTI \cite{geiger2012we} (left) and the PCV Kinect dataset \cite{MultiNormal2019} (right). The normal orientations are color-coded. 
On the left side, the estimated normals generated by DeepFit and PointNorm-Net are illustrated and zoomed in the blue and red boxes respectively for quality comparison. 
In contrast to the SOTA deep learning based approach DeepFit~\cite{DeepFit}, PointNorm-Net can better preserve sharp features and tiny structures while eliminating scanning noise more effectively. On the right side, the accuracy versus efficiency plot and 4 results on the PCV dataset indicate that PointNorm-Net is as fast as supervised deep methods, yet achieves much better accuracy.}
\vspace{-0.3cm} 
\label{fig:teaser}
\end{figure*}

An illustration can be found in \fig~\ref{fig:teaser}, where the point clouds of LiDAR and Kinect datasets present significantly different characteristics. It is not easy to accurately annotate point-wise normals for the noisy data from various sources, which makes them \jz{less} accessible for supervised learning. 
Conventional normal estimators \cite{MultiNormal2019, HF-cubes, least_squares, Zhang13, Bao, Jets, HoppeDDMS92} are data-independent. However it is widely known that they usually generate over-smoothed results \cite{MultiNormal2019, HF-cubes, Bao, DeepFit, Co-supported20} or required a long computation time, as illustrated in \fig~\ref{fig:teaser}. 
It is essential to design a more powerful and general normal estimation method for raw point clouds scanned by different 3D scanners without having to rely on costly normal annotations. 


This paper proposes PointNorm-Net, the first self-supervised deep method, for normal estimation. PointNorm-Net is designed based on a ``ground-truth sampling'' property that we established. This property states that given a smooth underlying surface and zero-mean noise, the ground-truth normal of a query point is the expectation of a set of randomly sampled candidate normals. 
Extensive observations of real-world point clouds indicate that sampled normals in most regions follow approximately unimodal distributions and meet the conditions necessary for the ground-truth sampling property. In such areas, a simple voting scheme can reliably estimate an accurate normal for a query point. However, this scheme may falter in areas with sharp features, where the distributions of sampled normals tend to be multimodal ~\cite{MultiNormal2019}. 
In these cases, accurately predicting normals, \ie~ estimating the major mode, requires a carefully designed approach.

The proposed PointNorm-Net combines a patch-based deep normal predictor with a training scheme empowered by an effective three-stage multimodal distribution estimation paradigm. The predictor could be any typical patch-based normal estimation network such as PCPNet~\cite{PCPNet:2018} and DeepFit~\cite{DeepFit}. 
Based on a multi-sample consensus scheme defined for local surface regions, the proposed three-stage multi-modal distribution estimation paradigm offers the predictor feasible candidate normals and training losses. These losses make the network learn the major mode of the distribution from numerous similar patches in training data, endowing the method with better generalization and scalability. During inference, there is only a forward pass of the network, thus ensuring efficiency.


PointNorm-Net avoids the need for costly annotations while retaining the efficiency and effectiveness of supervised methods. 
As shown in \fig~\ref{fig:teaser} and verified by comprehensive experiments, it outperforms both conventional methods and the cutting-edge deep learning methods on most open-source real-world datasets including Kinect \cite{MultiNormal2019,nyuv22012} and LiDAR datasets \cite{geiger2012we}. 
It is also worth noting that PointNorm-Net has competitive performance compared with supervised deep normal estimators, like HoughCNN and PCPNet, on most popular synthetic datasets.
Furthermore, the proposed multi-modal distribution estimation paradigm can be applied to other unsupervised point cloud processing problems as well. In Section \ref{sec:extension}, we demonstrate how to apply it to improve DMR~\cite{luo2020differentiable}, a state-of-the-art unsupervised point cloud denoising network.

The contributions of our work are summarized as follows:
\begin{itemize}
  \item We propose the first self-supervised deep learning normal prediction framework for 3D point cloud processing, which can be utilized in the large model training of 3D point clouds.  
  \item We propose an effective three-stage multi-modal distribution estimation paradigm, which is empowered by a local multi-sample consensus scheme for point clouds. 
This paradigm may improve deep learning methods for low-level point cloud processing, offering enhanced generalization and scalability.

  \item The proposed approach, PointNorm-Net, outperforms traditional optimization-based methods and supervised deep learning models on real-world Kinect and LiDAR point clouds.
\end{itemize}
\section{Related work}

\textbf{Optimization based normal estimators.} Normal estimation of point clouds is a long-standing problem in geometry processing, mainly because it is directly used in many downstream tasks. The most popular and simplest method for normal estimation is based on Principal Component Analysis (PCA) \cite{HoppeDDMS92}, which is utilized to find the eigenvectors of the covariance matrix constructed by neighbor points. Following this work, many variants have been proposed \cite{MLS,Estimating03,variants}. 
There are also methods based on Voronoi cells \cite{Nina,DeyG06,Pierre,Quentin}. But none of them are designed to handle outliers or sharp features \cite{Bao}. 
Sparse representation methods \cite{Zhang13,least_squares,NormalPei2014} and robust statistical techniques \cite{Bao,FleishmanCS05,Yoon,Mederos03,Wang2013,Wang2013Consolidation} are employed to improve normal estimation for point clouds with sharp features. 
And some other methods, such as Hough Transform \cite{HF-cubes}, also show impressive results. 
Although these algorithms are unsupervised and have strong theoretical guarantees, they are sensitive to noise. In addition, advanced traditional methods are usually much slower than deep normal estimators, since the latter infers normals only with a forward propagation.

\textbf{Deep network based normal estimators.} 
In recent years, the wide application and remarkable
success of deep learning have led to the proposal of normal estimation methods based on it. 
There have been some attempts \cite{Boulch2016,Roveri2018,NestiNet2019} to project local unstructured points into a regular domain, 
enabling the direct application of CNN architectures.
Along with the progress of geometric deep learning, some research endeavors to learn normals from irregular point clouds.
These methods can be roughly categorized into \textbf{Point Normal Learning networks} and \textbf{Point Weight Learning networks}. 
The Point Normal Learning network utilizes PointNet \cite{PCPNet:2018,Zhou2019,HashimotoS19} or graph neural networks \cite{PistilliFVM20} to extract a global feature that characterizes the local neighborhood and then learns a direct mapping from this global feature to the ground-truth normal.
On the other hand, the Point Weight Learning network employs a network to learn point-wise weights and obtains the normal vector through weighted least-squares polynomial surface fitting~\cite{DeepIterative,DeepFit, Adafit,Geometry-Guided,li2022graphfit}\jz{\cite{Rethinking23}}.
\jj{Recent advances in supervised methods have further pushed the boundaries of normal estimation accuracy on the synthetic dataset of PCPNet \cite{CMG24,MSECNet2024,SHS23}. HSurf-Net \cite{lihsurf} and SHS-Net \cite{SHS23} perform plane fitting in hyper spaces transformed from point clouds.  
NeAF \cite{li2023NeAF} infers an angle field of the input patch.
MSECNet \cite{MSECNet2024} improves estimation in normal varying regions by edge detection and edge conditioning. 
Nevertheless all the supervised deep normal estimators are trained using synthesized datasets.} 
Consequently, their performance usually deteriorates on real world datasets because of the domain gap between synthetic and real data.
Concurrent to our work, \cite{NeuralGF23} proposes unsupervised normal estimation by learning neural gradient functions. 
This approach outperforms previous unsupervised methods on two synthetic datasets, PCPNet \cite{PCPNet:2018} and FamousShape \cite{SHS23}, when optimized hyperparameters are used for each. However, fitting a global neural field to individual point clouds is computationally intensive, and the method's performance declines when applied to real datasets.

\textbf{Semi-supervised learning and Self-supervised learning.} 
Semi-supervised and Self-supervised learning are proposed to mitigate the reliance on manual annotations. 
Pseudo-labeling is a common semi-supervised learning strategy employed in advanced semantic tasks \cite{lee2013pseudo, berthelot2019mixmatch}. 
A pseudo label is typically generated for each sample using a pre-trained model. In contrast, our algorithm directly analyzes the point cloud to generate a set of low-cost pseudo labels for each point, eliminating the need for pre-training. It then identifies the major mode of the pseudo label distribution, rather than relying on individual labels.

Many low-level tasks, such as image denoising and point cloud denoising, also do not rely on manual annotations. They assume that noisy observations are stochastic realizations distributed around clean values. 
Lehtinen \etal~\cite{lehtinen2018noise2noise} utilize multiple noisy observations from image pairs to denoise an image.
Going one step further, Noise2Void \cite{krull2019noise2void} and Noise2Self \cite{batson2019noise2self} remove the requirement of paired noise corruptions and instead work on a single image.
Hermosilla \etal~\cite{totaldenoise} extend unsupervised image denoisers to 3D point clouds by introducing a proximity-appearance prior. The prior restricts the distribution of the candidate positions to have a single modal. 
Based on the core idea of \cite{totaldenoise}, Luo \etal~ \cite{luo2020differentiable} present an autoencoder-like neural network to improve the accuracy of denoising. 
Unfortunately, they cannot be naively extended to normal estimation, since they denoise the noisy observations, while we estimate the indirect geometry property of the noisy observations. They take an approximated expectation of multiple random samples as the denoised value. In contrast, we identify the major mode of candidate normals without relying on additional properties. 

\section{Overview}
The learning framework of PointNorm-Net is illustrated in \fig~\ref{fig:flow}.
The crucial technical component is a
learning paradigm composed of the following three stages: 1) initialization of candidate normals, 2) filtering of candidates normals, and 3) estimation of the major mode.
Given a 3D point cloud $P = \{\mathbf{p}_{t}~|~t = 1, 2, \cdots, N\}$ and a query point $\mathbf{p}_{t}$, PointNorm-Net first initializes a set of candidate normals $\dot{N}_{t}=\{\dot{\mathbf{n}}^{t}_{\theta}\}$ from its adaptive neighborhood $\mathcal{N}_{\hat{k}}$ (Stage~1 in \fig~\ref{fig:flow}). 
Then, infeasible normals are filtered out to form feasible candidates $\{\ddot{\mathbf{n}}^{t}_{\theta}\}$ (Stage~2 in \fig~\ref{fig:flow}). 
Finally, it utilizes a new loss function known as the candidate consensus loss function $L_{ccn}$, which is constructed based on feasible candidates $\{\ddot{\mathbf{n}}^{t}_{\theta}\}$, to optimize the predicted normal $\hat{\mathbf{n}}_{t}$ (Stage~3 in \fig~\ref{fig:flow}).
$\hat{\mathbf{n}}_{t}$ is expected to be the major mode of the distribution of the candidate normals, i.e. the feature-preserving normal with the greatest consensus among the feasible candidates.

During the stage of candidate normal initialization, candidate normals are randomly sampled from the neighborhood of the query point $\mathbf{p}_{t}$. 
For smooth surfaces where candidate normals follow \jz{a} unimodal distribution, we prove that the correct surface normal can be obtained by minimizing \eq~\ref{eq:op_2}, as detailed in subsection ~\ref{sec:Gen_candidate_normals}. 
However, when $\mathbf{p}_{t}$ is near sharp features, the underlying surface of its neighborhood usually consists of multiple piecewise smooth surfaces (corresponding to a multimodal distribution). In such areas, the correct normal can not be approximated by minimizing \eq~\ref{eq:op_2}.

\begin{wrapfigure}{}{0.41\columnwidth}
\vspace{-0.4cm} 
\includegraphics[width=1.0\linewidth]{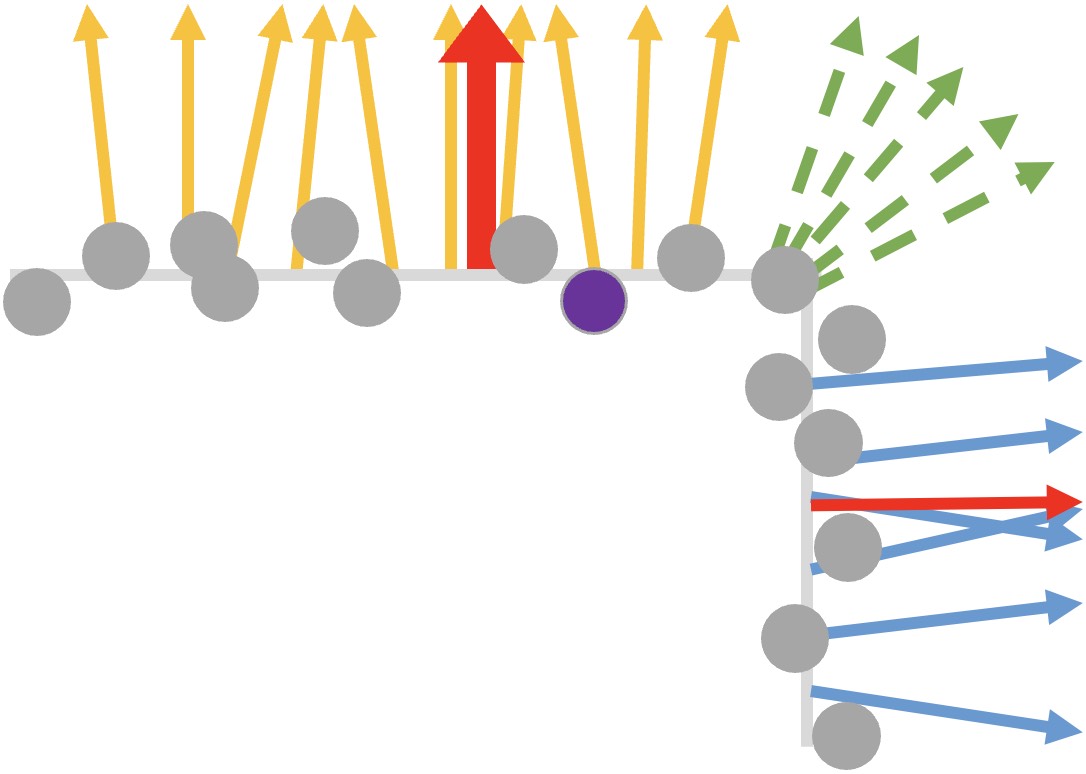} 
\caption{A 2D illustration of candidate normal distribution of a point close to a sharp edge.} 
\vspace{-0.35cm} 
\label{fig:wrapfig}
\end{wrapfigure}

As depicted in \fig~\ref{fig:wrapfig},
the randomly initialized candidates also contain some disturbing normals shown in blue, which are associated with the plane on the opposite side of the edge. Additionally, there are some blurry normals (colored green) if the neighbor points across the edge. 
Robust statistical techniques driven by multi-sample consensus are employed to eliminate the influence of these unwanted candidate normals. The blurry normals can be filtered out through the consensus of the neighbor points of $\mathbf{p}_{t}$, \ie~there are only a few neighbor points supporting a plane with a blurry normal (subsection \ref{sec:Sel_candi_normasls}).
 Among the remaining feasible candidate normals, the disturbing modes have fewer supporters. The major mode of the feasible candidates has the greatest consensus among them.
We design a new candidate consensus loss (subsection \ref{sec:normal_loss}) to estimate the major mode that corresponds to the correct normal. 
 The network and other auxiliary losses will be introduced in subsection ~\ref{sec:netandloss}.

\section{PointNorm-Net}\label{sec:PointNorm}
\subsection{{Multimodal Normal Distribution Estimation}}\label{sec:MCRNE}
\subsubsection{Candidate normal initialization} \label{sec:Gen_candidate_normals}

\begin{figure*}
\begin{center}
\begin{tabular}{@{}c@{} }
\includegraphics[width=1\textwidth, keepaspectratio]{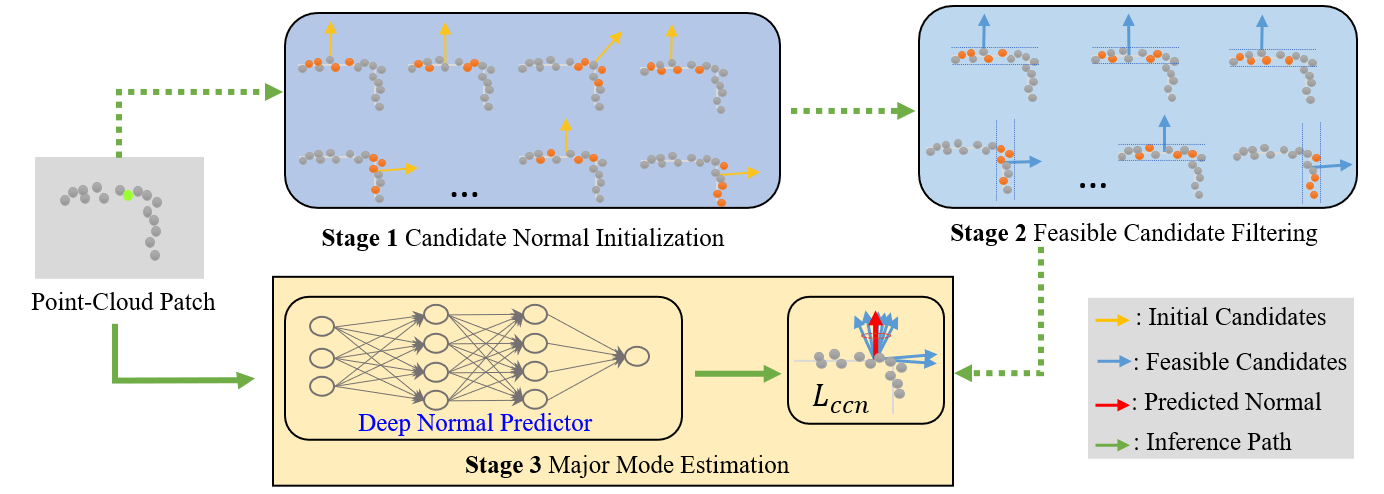}
\end{tabular}
\end{center}
   \caption{The training pipeline of the proposed PointNorm-Net. PointNorm-Net is a self-supervised and network-agnostic framework. Many patch-based normal estimation networks can be employed for the normal predictor. Note that during inference, the network only requires a single forward pass (\ie, the stage of major mode estimation), excluding stages 1 and 2.}
\vspace{-0.5cm} 
\label{fig:flow}
\end{figure*}

The initial candidate normals $\dot{N}_{t}=\{\dot{\mathbf{n}}^{t}_{\theta}\}$ of the query point are built from its adaptive neighborhood $\mathcal{N}_{\hat{k}}(\mathbf{p}_t)$ of size $\hat{k}$. \jz{Specifically, we first randomly select $k_s$ points from $\mathcal{N}_{\hat{k}}(\mathbf{p}_t)$. Then, we fit a plane $\theta$ to these points using the least-squares method and define the normal vector of the plane as $\dot{\mathbf{n}}^{t}_{\theta}$. By repeating this process multiple times, we can efficiently construct a large set of candidates for each query point.
When $k_s = 3$, it} is proven in the following that the noise-free normal $\mathbf{n}_{t}$ is the expectation of $\dot{\mathbf{n}}^{t}_{\theta}$ if the underlying surface of the noisy point cloud is smooth and the expectation of the noise is zero. 
Hence a faithful approximation can be achieved if $\dot{N}_{t}$ includes sufficient candidates.

\textbf{Theoretical analysis.}
Let $\tilde{\mathbf{p}}_{t}$ be a point on the underlying smooth surface $S$ of the point cloud. $U$ is an open set on $S$ centering at $\tilde{\mathbf{p}}_{t}$. The normal of it is denoted as $\mathbf n_{t}$. For any three non collinear points $\{\tilde{\mathbf{p}}^{t}_{\theta 1}, \tilde{\mathbf{p}}^{t}_{\theta 2}, \tilde{\mathbf{p}}^{t}_{\theta 3}\}$ on U, $\tilde{\mathbf{n}}^{t}_{\theta}=\left(\tilde{\mathbf{p}}^{t}_{\theta 2}-\tilde{\mathbf{p}}^{t}_{\theta 1}\right) \times\left(\tilde{\mathbf{p}}^{t}_{\theta 3}-\tilde{\mathbf{p}}^{t}_{\theta 1}\right)$ is the vector corresponding to the plane $\theta$ spanned by $\{\tilde{\mathbf{p}}^{t}_{\theta 1}, \tilde{\mathbf{p}}^{t}_{\theta 2}, \tilde{\mathbf{p}}^{t}_{\theta 3}\}$.
Their noisy observations are $\left\{\mathbf{p}^{t}_{\theta 1}, \mathbf{p}^{t}_{\theta 2}, \mathbf{p}^{t}_{\theta 3}\right\}$, where $\mathbf{p}^{t}_{\theta j}=\tilde{\mathbf{p}}^{t}_{\theta j}+\mathbf{\varepsilon}_{j}, j=1,2,3$.
Each group of them defines an candidate normal  $\dot{\mathbf{n}}^{t}_{\theta}$. 
Here, $\{\tilde{\mathbf{p}}^{t}_{\theta 1}, \tilde{\mathbf{p}}^{t}_{\theta 2}, \tilde{\mathbf{p}}^{t}_{\theta 3}\}$ are sorted counterclockwise and noise $\mathbf{\varepsilon}_{j}=(\varepsilon_{xj},\varepsilon_{yj},\varepsilon_{zj})^{T}$ is assumed to be independent for components. For convenience, we do not normalize $\tilde{\mathbf{n}}^{t}_{\theta}$ and $\dot{\mathbf{n}}^{t}_{\theta}$ in the following theoretical analysis. 
Then we can obtain the following Theorem and Corollary. Their proofs are shown in the supplementary material.

\begin{theorem}\label{th:2}
If $E\{\varepsilon_{j}\}=0$, then we have $E\left\{\dot{\mathbf{n}}^{t}_{\theta}\right\}=E\left\{\tilde{\mathbf{n}}^{t}_{\theta}\right\}$. 
\end{theorem}

\begin{corollary}
If $S$ is smooth, $U$ is a small enough neighborhood and $E\{ \varepsilon\}=0$, then $E\{\dot{\mathbf{n}}^{t}_{\theta}\}$ is on the same line with $\mathbf n_{t}$.
\end{corollary}


The minimum of the following optimization problem
\begin{equation}\label{eq:op_1}
\underset{\mathbf{z}}{\operatorname{argmin}} E_{\dot{\mathbf{n}}^{t}_{\theta}}\left\{\|\mathbf{z}-\dot{\mathbf{n}}^{t}_{\theta}\|_{F}^{2}\right\},
\end{equation}
is found at the expectation of $\dot{\mathbf{n}}^{t}_{\theta}$, \ie~$\mathbf{z}=E\{\dot{\mathbf{n}}^{t}_{\theta}\}$. 
We have proven that $E\{\dot{\mathbf{n}}^{t}_{\theta}\}=k \mathbf{n}_{t}$.
Therefore, given finite candidate normals $\dot{N}_{t}$, we can approximate the direction of $\mathbf n_{t}$ by the following optimization problem
\begin{equation}\label{eq:op_2}
\underset{\mathbf{z}}{\operatorname{argmin}}\sum_{\dot{\mathbf{n}}^{t}_{\theta}\in \dot{N}_{t}}\left\{\|\mathbf{z}- \dot{\mathbf{n}}^{t}_{\theta}\|_{F}^{2}\right\}, 
\end{equation}
when the surface is smooth. In the implementation, the candidate normals $\dot{\mathbf{n}}^{t}_{\theta}$ are normalized. To avoid introducing unnecessary symbols, we still use $\dot{\mathbf{n}}^{t}_{\theta}$ to represent the normalized vector in the following discussion. 
It is noteworthy that the larger the size of the candidate set $\dot{N}_{t}$, the better the approximation effect.

\begin{figure}
\setlength{\abovecaptionskip}{-0.1cm}
\begin{center}
\begin{tabular}{@{}c@{} }
\includegraphics[width=0.5\textwidth, keepaspectratio]{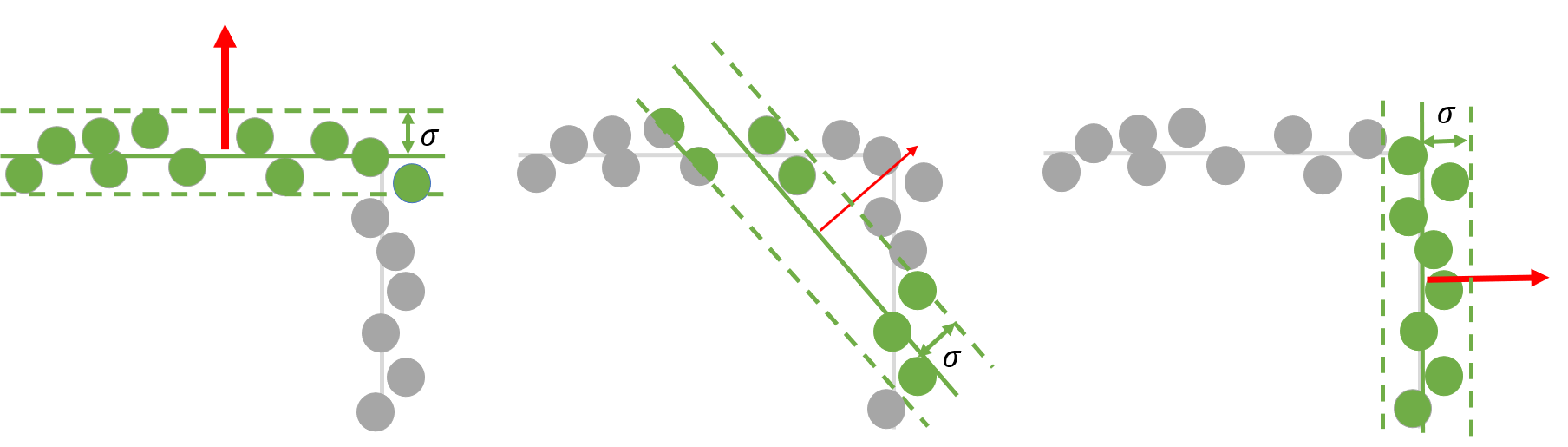}
\end{tabular}
\end{center}
\caption{\re{We expect to find normals corresponding to planes with more inlier points (green points). The thickness of the normal is determined by the number of inlier points.}}
\vspace{-0.5cm} 
\label{fig:selection}
\end{figure}

\textbf{Adaptive neighborhood} \label{sec:Optimization_NeighborSize}
In theoretical analysis, the neighborhood $\mathcal{N}_{\hat{k}}(\mathbf{p}_t)$ should be as small as possible to be regarded as a plane. However, a small neighborhood is inadequate to depict the underlying structure of the surface when the point cloud is polluted by large noise. Therefore, the neighborhood size is adaptively determined. Specifically, for each point $\mathbf{p}_{t}$, we ﬁrst apply the local covariance analysis to characterize its noise level $f_{t}$ (Section 4 in  \cite{MultiNormal2019}). The average $f=\sum_{t=1}^{N}f_{t}/N$ is then utilized to describe the noise scale of the whole point cloud. Finally, a suitable neighborhood size is determined according to the interval in which $f$ is located: $\hat{k} = k_{i}, if~l_{i-1} \le f< l_{i}$. We take $k_{1}=32$, $k_{2}=128$, $k_{3} = 256$, $k_{4}=450$, $l_0=0$, $l_1=0.02$, $l_2=0.14$, and $l_3=0.16$, $l_4=0.3$ in our experiments.

\subsubsection{Feasible candidate filtering}
\label{sec:Sel_candi_normasls}

The randomly initialized candidates are inevitably noisy. Drawing inspiration from kernel density estimation, we first filtered out the blurry normals by the consensus among the neighbors of $\mathbf{p}_{t}$. For each candidate normal $\dot{\mathbf{n}}^{t}_{\theta}$, we define a score $s^{t}_{\theta}$ based on the distances of all neighbor points to its corresponding plane $\theta$:
\begin{equation} \label{eq:FSort}
s^{t}_{\theta}=\sum_{i=1}^{\hat{k}} e^{-\frac{\left(d(\mathbf{p}_{t}^{i},\theta)\right)^{2}}{\sigma^{2}}}, 
\end{equation}
where $\mathbf{p}_{t}^{i} \in \mathcal{N}_{\hat{k}}(p_t)$ is a neighbor point of $\mathbf{p}_{t}$, $d(\mathbf{p}_{t}^{i},\theta)$ represents the Euclidean distance from point $\mathbf{p}_{t}^{i}$ to the plane $\theta$, and $\sigma$ is the bandwidth of the Gaussian kernel function.
A higher score means a higher consensus reached by the query point's neighbor points. It is designed to filter out normals supported by fewer inliers, as shown in \fig~\ref{fig:selection}. 
When $d(\mathbf{p}_{t}^{i},\theta)$ is less than the bandwidth $\sigma$, the point $\mathbf{p}_{t}^{i}$ is regarded as an inlier point of the plane $\theta$. It will give $\theta$ a strong support $e^{-\left(d(\mathbf{p}_{t}^{i},\theta)\right)^{2}/\sigma^{2}}$. The larger $s^{t}_{\theta}$ is, the more feasible the normal is. As a result, we sort the initialized candidate normals based on their scores and remove the lowest 10\% of candidates.
The remaining normals are feasible candidate normals, denoted by $\ddot{N}_{t}=\{\ddot{\mathbf{n}}^{t}_{\theta}\}$. 

The value of $\sigma$ should depend on the noise level of the point cloud. It seems that $\sigma$ should take a large value for point clouds with large noise since a high noise level will result in more neighbor points with large $d(\mathbf{p}_{t}^{i},\theta)$ even the plane is feasible. Conversely, when the noise of the point cloud is small, $\sigma$ should take a smaller value. 
However, in experiments, we found that a large $\sigma$ does not work well for either large or small noise. Therefore, we only employ this filtering strategy for noise-free or small noise point clouds whose average noise level $f$ is located in the first two intervals, as stated in Section \ref{sec:Optimization_NeighborSize}. They usually require a small $\sigma$. We set it as one percent of the neighborhood radius.

\subsubsection{Major mode estimation}
\label{sec:normal_loss}
If $\mathbf{p}_{t}$ lies near some sharp features, it is difficult to accurately exclude the interference of disturbing normals by the score defined by Eq. \eqref{eq:FSort}. The distribution of $\ddot{N}_{t}$ is still possibly a multi-modal distribution, as demonstrated in \fig~\ref{fig:wrapfig}. We formulate the problem as follows.

\textbf{Problem formulation:} Given the candidate normal set $\{\mathbf{n}_1,\mathbf{n}_2,\dots,\mathbf{n}_m\}$ of a point, we wish to find a subset $S$ from the set. The variation of normals in $S$ is as small as possible and the number of normals is as large as possible, that is:
\begin{equation}\label{eq:problem}
\max\left \{ card(S) + \sum_{\mathbf{n}_i,\mathbf{n}_j\in S}cos(\mathbf{n}_{i},\mathbf{n}_{j}) \right \},
\end{equation} where $card(\cdot)$ represents the cardinality of a set. 


Inspired by robust statistical techniques, we treat this problem as a soft optimization problem. A candidate consensus loss function for normal estimation is proposed as follows.
\begin{equation}\label{eq:normal_loss}
L_{ccn}=\sum_{\theta}^{}-e^{-\frac{\left\|\hat{\mathbf{n}}_{t} \times \ddot{\mathbf{n}}^{t}_{\theta}\right\|_{F}^{2}}{\tau^{2}}}, ~~~ \ddot{\mathbf{n}}^{t}_{\theta}\in \ddot{N}_{t}, 
\end{equation}
where $\hat{\mathbf{n}}_{t}$ is the predicted normal, $\tau=\sin\alpha$ denotes the bandwidth of the Gaussian kernel function. $\alpha=\frac{\pi}{6}$ in all the experiments.

\begin{figure}
\setlength{\abovecaptionskip}{-0.2cm}
\begin{center}
\begin{tabular}{@{}c@{} }
\includegraphics[width=0.4\textwidth, keepaspectratio]{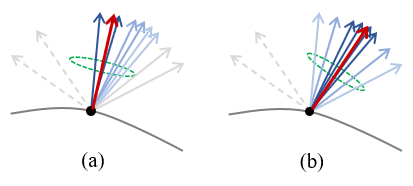}
\end{tabular}
\end{center}  
\caption{Candidate consensus loss function can effectively exclude the influence of disturbing normals (dotted normals). The red normals in (a) and (b) are the average normals of all candidates and solid candidates, respectively. The candidates in the green circles are regarded as inliers of the red normals. The darker the color, the higher their contribution to the red normals, which are determined by the candidate consensus loss.}
\vspace{-0.2cm} 
\label{fig:candidate_loss}
\end{figure}

When the angle between the predicted normal and the candidate normal is less than $\alpha$, we consider this candidate to be an inlier of the predicted normal. Thus a predicted normal with the most inliers, \ie~ with the greatest consensus of the candidate normals, is desired. As shown in \fig~\ref{fig:candidate_loss}, this mechanism can effectively overcome the influence of disturbing normals.

\subsection{Network and losses}\label{sec:netandloss}
Patch-based learning is the most common approach adopted by existing deep normal estimators \cite{PCPNet:2018, DeepFit, Geometry-Guided}. 
Generally, any patch-based normal learning network may be employed. We follow the architecture of DeepFit \cite{DeepFit}, which learns point-wise weight for weighted least squares fitting. Note that we use the 1-order jet instead of the 3-order jet which is the default setting of DeepFit since we estimate the local plane.
Specifically, the input $\mathcal{N}_k(p_t)$ is passed through PointNet to extract a feature for each point. Next, the feature is fed to the multi-layer perceptron (MLP) to output weight $w_{i}$ for each neighbor point $\mathbf{p}_{t}^{i}$ in $\mathcal{N}_k(p_t)$. Finally, these weights are used to fit a plane whose normal is the estimated normal of $\mathbf{p}_{t}$. More details are referred to Section 3.1 of \cite{DeepFit}. 
We further design a feature loss function for the estimated weights to recover the sharp features.

\begin{table*}[htbp]\small
\begin{center}
\caption{\re{Comparison of RMS, RMS$_\tau$, and PGP$\alpha$ on the Kinect dataset of PCV~\cite{MultiNormal2019}. Both PointNorm-Opt and PointNorm-Net surpass previous conventional methods and deep learning methods. (Sup.: Supervision, Unsup.: Unsupervision, Self-sup.: Self-supervision)}}
\vspace{-0.2cm} 
\label{tab:RMS_PCV_V1}
\begin{tabular}{m{0.16\textwidth}<{\raggedright} m{0.06\textwidth}<{\centering }|m{0.08\textwidth}<{\centering } m{0.08\textwidth}<{\centering } m{0.08\textwidth}<{\centering }|m{0.08\textwidth}<{\centering } m{0.08\textwidth}<{\centering } m{0.08\textwidth}<{\centering } m{0.08\textwidth}<{\centering }}
\hline
Models  &  RMS $\downarrow$  & RMS\_10 $\downarrow$ & RMS\_15 $\downarrow$ &RMS\_20 $\downarrow$ & PGP10 $\uparrow$ & PGP15 $\uparrow$ & PGP20 $\uparrow$ & PGP25 $\uparrow$  \\
\hline
\textbf{Sup.} & & & &  & \\
SHS-Net   & 16.13 &64.31 & 50.58 & 39.38 & 0.4905 & 0.6888 & 0.8181  & 0.8945 \\
CMG-Net   & 17.93 &67.77 & 54.79 & 43.69 & 0.4340 & 0.6339 & 0.7743  & 0.8630 \\
MSECNet  & 17.98 & 68.84 & 55.90 & 44.38 & 0.4153 & 0.6183 &0.7666  & 0.8617\\
AdaFit & 18.38 &66.25& 54.17 & 44.32 & 0.4593 & 0.6421 &0.7666 & 0.8495\\
G2N2   &17.57 &64.27 &51.40 & 41.22 &0.4909& 0.6782 & 0.7993 & 0.8743  \\
DeepFit & 13.68 & 58.01 & 43.71&32.81 & 0.5853 & 0.7682 & 0.8753 &0.9343\\
PCPNet  & 15.73 &62.73 &47.17 & 35.62 & 0.5155& 0.7302 &0.8527 & 0.9162 \\
\hline  
\textbf{Unsup. $\&$ Self-sup. } & & & &  & \\
HF-cubes  &11.72& 43.99 & 31.79 & 24.73 &0.7568 & 0.8747 & 0.9274 & 0.9536\\    
LRRfast   &13.25 & 48.36 &34.23 &26.84 &0.7103 & 0.8573 & 0.9162 & 0.9452\\ 
PCV       & 11.74 &46.85 &31.81& 23.85 & 0.7279 & 0.8768 &0.9347 & 0.9607\\
NeuralGF & 15.67 & 63.02 & 47.31 & 35.46 &0.5099  &0.7265  &0.8518 & 0.9183  \\
PointNorm-Opt     &11.67& 49.54& 33.64& 24.51&  0.6967 &     0.8628  &0.9316 &0.9620\\
PointNorm-Net  & \textbf{10.89} & \textbf{43.75} & \textbf{30.82} &\textbf{23.40} &\textbf{0.7614} & \textbf{0.8836} & \textbf{0.9363} & \textbf{0.9631} \\
\hline
\end{tabular}
\end{center}
\vspace{-0.3cm} 
\end{table*}

\subsubsection{Feature-preserving loss}
\label{sec:Edge_loss}
Two neighbor points with significantly different normals tend to be located on different surfaces divided by sharp features.
Therefore, these two points should not have large weights at the same time when fitting a plane. 
Based on this fact, we design a feature-preserving loss to constrain the point-wise weight learned by the network, which is beneficial to recovering sharp features.
The feature loss function is formulated as follows:
\begin{equation}\label{eq:loss_w}
L_{w}=\sum_{i,j=1}^{\hat{k}}e^{-\frac{\left\langle \mathbf{n}_{i}, \mathbf{n}_{j}\right\rangle^2}{\omega^{2}}} \cdot w_{i} \cdot w_{j}, 
\end{equation}
where $\mathbf{n}_{i}$ is the normal of $\mathbf{p}_{t}^{i}$ computed by PCA with neighborhood size $\hat{k}$, $w_{i}$ is the learned weight of $\mathbf{p}_{t}^{i}$, and $\omega=\cos\frac{\pi}{3}$ is a hyperparameter. 
When $n_{i}$ and $n_{j}$ are significantly different (angles between them larger than $\frac{\pi}{3}$), $e^{-\frac{\left\langle \mathbf{n}_{i}, \mathbf{n}_{j}\right\rangle^2}{\omega^{2}}}$ will be large. This will prevent $w_{i}$ and $w_{j}$ from achieving large values at the same time.

\subsubsection{Total loss}
\label{sec:Total}
The total loss to train the network includes four terms: the candidate consensus loss $L_{\text {ccn}}$ between the predicted normal and multi-candidate normals at the query point, the feature loss $L_{w}$ for weights learned by the network, the regularization loss $L_{\text{regw}}=-\frac{1}{k} \sum_{j=1}^{k} \log \left(w_{j}\right)$ preventing all of weights from being optimized to 0, and a transformation matrix regularization term $L_{\text{regm}}=\left|I-A A^{T}\right|$ where $I$ is an identity matrix and $A$ is the transformation matrix used in PointNet. On the whole, it is defined as follows: 
\begin{equation}
L_{\text {total }}= L_{\text {ccn}}+ \gamma \cdot L_{w} + \delta_1 \cdot L_{\text {regw}} + \delta_2 \cdot L_{\text {regm}}, 
\end{equation}
where $\gamma$, $\delta_1$ and $\delta_2$ are hyper-parameters used to balance these four terms. We set $\gamma=10^{-4}$, $\delta_1=5\times10^{-2}$ and $\delta_2=10^{-2}$ in all the experiments.

\begin{figure*}[htbp]
\begin{overpic}[width=1\textwidth, keepaspectratio]{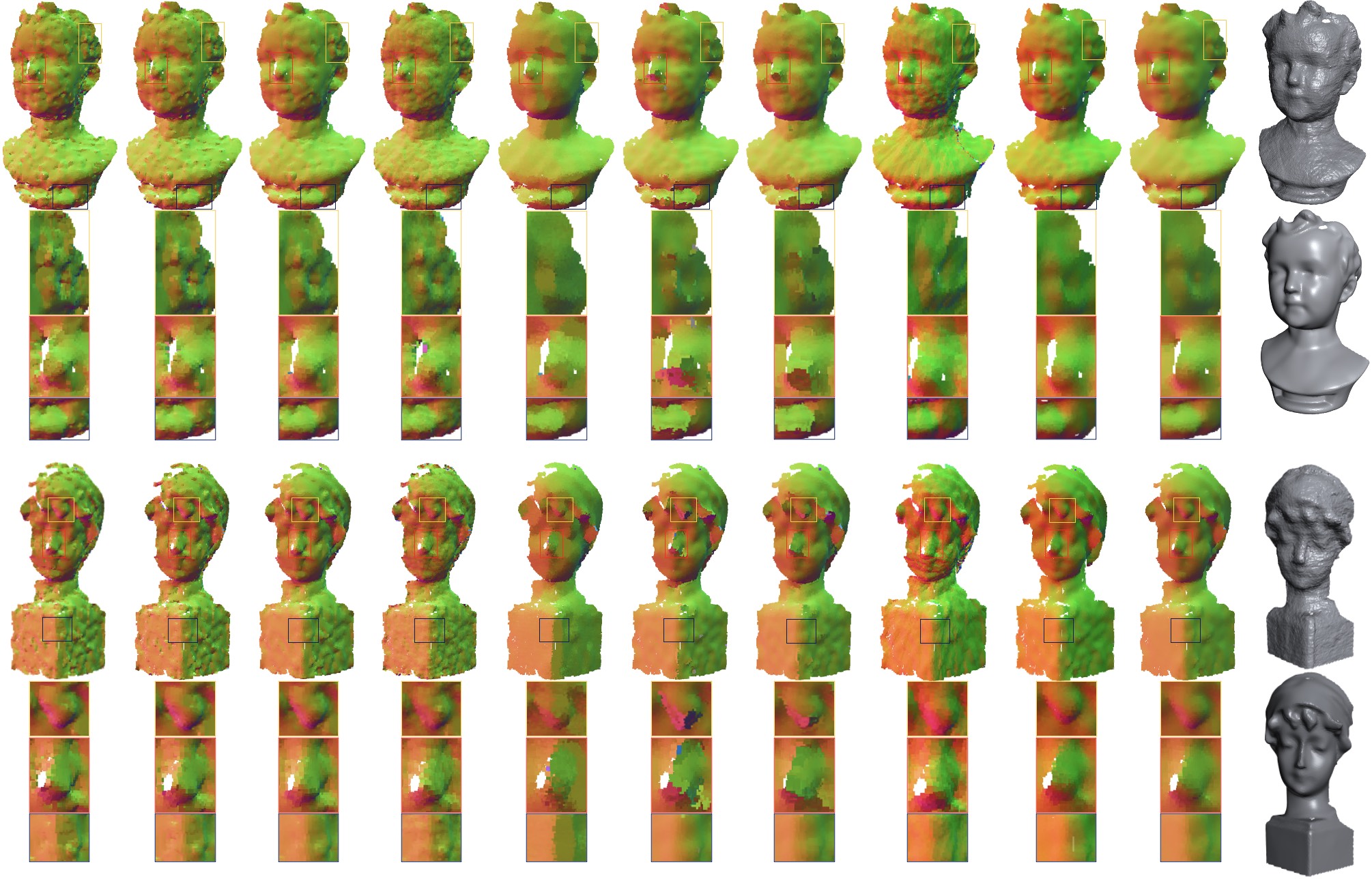}

\put(2.6,30.8){16.54}
\put(11.8,30.8){15.90}
\put(20.8,30.8){11.85}
\put(29.7,30.8){14.11}
\put(38.8,30.8){12.03}
\put(47.9,30.8){12.50}
\put(56.9,30.8){11.73}
\put(66.7,30.8){13.39}
\put(76,30.8){10.58}
\put(85,30.8){\textbf{10.31}}

\put(2.6,-0.2){20.90}
\put(11.8,-0.2){19.79}
\put(20.8,-0.2){15.85}
\put(29.7,-0.2){18.16}
\put(38.8,-0.2){15.47}
\put(47.9,-0.2){15.14}
\put(56.9,-0.2){14.50}
\put(66.7,-0.2){17.12}
\put(76,-0.2){14.04}
\put(85,-0.2){\textbf{13.45}}

\put(1.8,-2){\small AdaFit}
\put(11.3,-2){\small G2N2}
\put(19.6,-2){\small DeepFit}
\put(28.4,-2){\small PCPNet}
\put(37,-2){\small HF-cubes}
\put(46.9,-2){\small LRRfast}
\put(56.8,-2){\small PCV}
\put(64.7,-2){\small NeuralGF}
\put(73.4,-2){\small PointNorm}
\put(75.2,-3.6){\small -Opt}
\put(82.5,-2){\small PointNorm}
\put(84.3,-3.6){\small -Net}
\end{overpic}
\vspace{0.2cm} 
    \caption{Visual comparison of estimated normals on two statues scanned by Kinect from PCV~\cite{MultiNormal2019}. Supervised deep normal estimators suffer from the scanner noise, \ie~the small fluctuations in smooth regions. The conventional methods can handle the fluctuations but may introduce artifacts around sharp features. PointNorm-Net works well for both two cases. The statues scanned by a high-precision Artec Spider$^{TM}$ scanner and a Kinect are shown in the rightmost column respectively. The point normal vectors are mapped to RGB colors. Numbers show RMS of the results.}
\vspace{-0.5cm} 
\label{fig:PCV}
\end{figure*}
\begin{figure*}[htbp]
\vspace{-0.6cm} 
\setlength{\abovecaptionskip}{0cm}
\begin{center}
\begin{tabular}{@{}c@{} }
\includegraphics[width=1\textwidth, keepaspectratio]{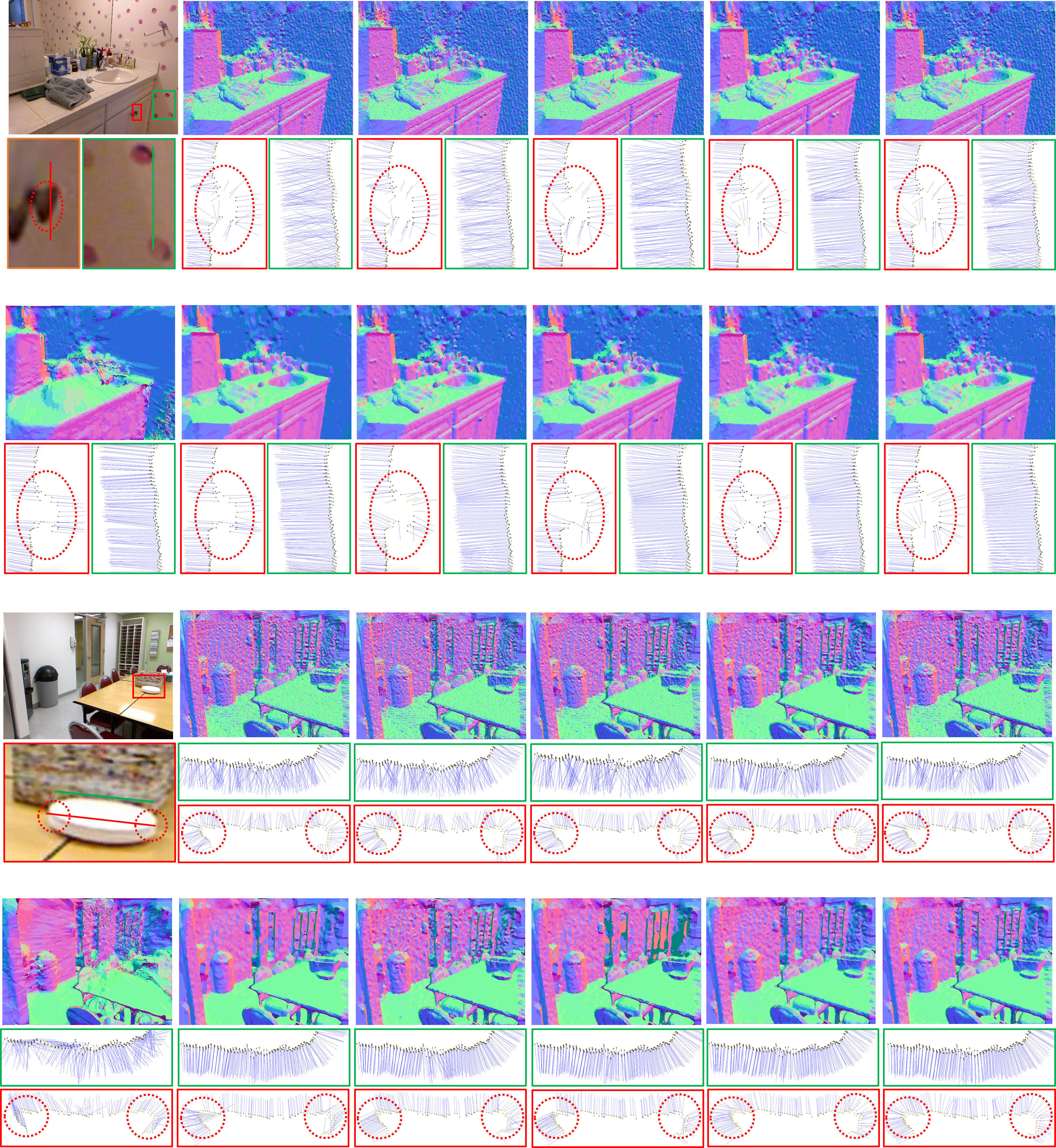}
\put(-490,420){Image}
\put(-410,420){MSECNet}
\put(-325,420){CMG-Net}
\put(-235,420){SHS-Net}
\put(-147,420){DeepFit}
\put(-58,420){AdaFit}

\put(-495,270){NeuralGF}
\put(-410,270){HF-cubes}
\put(-320,270){LRRfast}
\put(-226,270){PCV}
\put(-165,270){PointNorm-Opt}
\put(-75,270){PointNorm-Net}
\put(-490,128){Image}
\put(-410,128){MSECNet}
\put(-325,128){CMG-Net}
\put(-235,128){SHS-Net}
\put(-147,128){DeepFit}
\put(-58,128){AdaFit}

\put(-495,-10){NeuralGF}
\put(-410,-10){HF-cubes}
\put(-320,-10){LRRfast}
\put(-226,-10){PCV}
\put(-165,-10){PointNorm-Opt}
\put(-75,-10){PointNorm-Net}
\end{tabular}
\end{center}
\vspace{0.2cm}
\caption{\re{Visual comparison of estimated normals on two scanned scenes from NYUV2. After performing unoriented normal estimation, we flip normals according to the camera position. Colors encode the direction of oriented normals. SHS-Net, MSECNet, CMG-Net, AdaFit, and DeepFit can not suppress scanner noise, refer to the plane regions, such as walls and upper surfaces of tables. HF-cubes, LRRfast and PCV  may fail to preserve tiny structures. PointNorm-Net overcomes these challenges.}} 
\label{fig:nyuv2}
\vspace{-0.4cm} 
\end{figure*}
\input{figures/fig_kitti-1.tex}
\section{Experiments}
\label{sec:experiments}
We conducted comprehensive experiments using six datasets: four acquired from various 3D scanning technologies (KITTI \cite{geiger2012we}, Semantic3D \cite{Semantic3D}, NYUV2 \cite{nyuv22012}, and the dataset from PCV \cite{MultiNormal2019}) and two sampled from triangular meshes (SceneNN \cite{SceneNN} and the dataset from PCPNet \cite{PCPNet:2018}).  
The performance is evaluated with the \emph{Root Mean Square} (RMS) error of the angle difference \cite{PCPNet:2018}, \emph{RMS with threshold} (RMS$_\tau$) which makes the measure greater than $\tau^{\circ}$ as bad as $90^{\circ}$ \cite{HF-cubes}, and the \emph{Percentage of Good Points} (PGP$\alpha$) metric which computes the percentage of points with an error less than $\alpha$ degree \cite{PCPNet:2018}. 

To thoroughly evaluate the performance of our approach, two categories of methods are compared: 1) unsupervised methods including traditional HF-cubes \cite{HF-cubes}, LRRfast \cite{least_squares}, PCV \cite{MultiNormal2019}, PCA \cite{HoppeDDMS92}, and learning-based NeuralGF \cite{NeuralGF23}; 2) supervised deep learning methods including \jz{SHS-Net \cite{SHS23}, CMG-Net \cite{CMG24}, MSECNet \cite{MSECNet2024},} AdaFit \cite{Adafit}, G2N2 \cite{Geometry-Guided}, DeepFit \cite{DeepFit}, and PCPNet \cite{PCPNet:2018}. 
The same scale with 256 neighbor points is chosen for the traditional methods. All other parameters, if any, are set to default. The parameters of NeuralGF for SceneNN, Semantic3D, KITTI, and the synthetic dataset of PCPNet are set as the authors provided and tuned to obtain the best results for PCV and NYUV2. For the supervised deep learning methods, we employed the original configurations as specified by their respective authors.

For PointNorm-Net, we set the following parameters: input neighborhood size ($k$) to 256, number of randomly selected points for fitting candidate normals ($k_s$) to 4, and number of initial candidate normals to 100. We use Adam optimizer with a batch size of 512 and a learning rate of 0.001. The implementation is done in PyTorch and trained on a single Nvidia GTX 1080Ti.

\subsection{A baseline method}
To comprehensively evaluate the effectiveness of the multi-modal normal distribution estimation paradigm, we also propose an optimization-based method and name it PointNorm-Opt as a baseline. Given the input patch $\mathcal{N}_k(p_t)$ and the query point $\mathbf{p}_{t}$, PointNorm-Opt employs a pipeline consisting of the stages 1 and 2 in PointNorm-Net. 
Without using any network based normal predictor, PointNorm-Opt estimates the normal of point $\mathbf{p}_{t}$ by directly minimizing the candidate consensus loss function:
\begin{equation}\label{eq:8}
\hat{\mathbf{n}}_{t}=argmin_{\hat{\mathbf{n}}_{t}}\sum_{\theta}^{}-e^{-\frac{\left\|\hat{\mathbf{n}}_{t} \times \ddot{\mathbf{n}}^{t}_{\theta}\right\|_{F}^{2}}{\tau^{2}}}, ~~~ \ddot{\mathbf{n}}^{t}_{\theta}\in \ddot{N}_{t}, 
\end{equation}
where $\hat{\mathbf{n}}_{t}$ is the estimated normal. In our comparison experiments, PointNorm-Opt adopts the same hyper-parameters as PointNorm-Net.
\subsection{Evaluation on real-world scanned point clouds}
In this section, four real scanned datasets, including two Kinect datasets \cite{nyuv22012, MultiNormal2019}, one LiDAR dataset \cite{geiger2012we}, and one terrestrial laser scanner dataset \cite{Semantic3D}, are utilized to evaluate our methods. All supervised learning-based methods were trained on the PCPNet dataset and directly evaluated on these real-world scanned datasets, as actual scan datasets typically lack ground truth normals. To ensure a fair comparison, we trained our PointNorm-Net solely on the synthetic PCPNet dataset.
NeuralGF learns a global neural field from each point cloud, hence it has to be trained on each data.

\subsubsection{Evaluation Kinect datasets}
We begin by evaluating the performance of our methods using two real datasets captured by a Kinect v1 RGBD camera.  
These data reveal additional challenges, such as fluctuations on flat surfaces, originating from the projection process of the Kinect camera. The non-Gaussian and discontinuous noise in the scanned data results in distinct noise patterns compared to synthetic data. Specifically, the noise often matches the magnitude of tiny features, making it difficult to distinguish between the two \cite{DeepFit,Co-supported20}, see \fig~\ref{fig:PCV} and \ref{fig:nyuv2}. It is observed that our PointNorm-Net demonstrates better generalization ability. It can effectively address the imperfections of these data, even when training only on synthetic data polluted by Gaussian noise.

\textbf{Dataset from PCV ~\cite{MultiNormal2019}.} There are 71 scans in the training set and 73 scans in the test set. For each scan, a registered and reconstructed mesh from an Artec Spider$\mathrm{^{TM}}$ scanner (accuracy 0.5 mm) is used to build ground truth normals. 
\tab~\ref{tab:RMS_PCV_V1} shows RMS, RMS$_\tau$, and PGP$\alpha$ for the test set. Some visual comparisons of estimated normals rendered in RGB color are shown in \fig~\ref{fig:PCV}. From \tab~\ref{tab:RMS_PCV_V1}, we see that our methods achieve the lowest RMS error and the highest PGP metric. Furthermore, in contrast to the findings from the synthetic dataset, our comparison reveals that all supervised methods underperform relative to the traditional methods. Although \jz{some recent approaches, such as SHS-Net, CMG-Net and MSECNet} exhibit improved performance over DeepFit on synthetic datasets, they demonstrate weaker generalization capabilities in comparison to DeepFit. This phenomenon occurs primarily because the noise patterns in synthetic datasets differ from those in real scanned datasets. The supervised methods misidentify certain noises as features, which disrupt the structure of smooth regions, as demonstrated in the results of AdaFit, G2N2, DeepFit, and PCPNet in \fig~\ref{fig:PCV}. This underscores the necessity of fundamentally enhancing the generalization ability of supervised deep normal estimators.

Unsupervised NeuralGF also performs poorly on the PCV dataset, see \tab~\ref{tab:RMS_PCV_V1}. It even introduces an obvious artifact on the shoulder of the statues in the top row of \fig~\ref{fig:PCV}. 
The traditional methods, including HF-cubes, LRRfast, and PCV, handle the fluctuation introduced by the scanner better than supervised methods. However, they tend to trigger off artifacts around sharp features, such as the noses of the two statues in \fig~\ref{fig:PCV}. Our PointNorm-Net and PointNorm-Opt work well for both cases. They can better trade off features and smooth regions compared with other methods. Compared with PointNorm-Opt, PointNorm-Net is more robust. 
It performs comparably to PointNorm-Opt in terms of feature preservation, yet the estimated normals are smoother and exhibit reduced sensitivity to noise. 

\textbf{NYUV2 dataset.} We also evaluate the performance of the proposed methods on the NYUV2 dataset \cite{nyuv22012}. The dataset consists of indoor scenes without ground truth normals, so we only perform a qualitative comparison of various methods. Due to the page space limitation, we only show the visual comparison of estimated normals on two scanned scenes. More results are presented in the supplementary material. 
As shown in \fig~\ref{fig:nyuv2}, \jz{SHS-Net, CMG-Net, MSECNet, AdaFit, and DeepFit} can preserve tiny details but with the price of retaining scanner noise, refer to the plane regions, \re{such as walls and upper surfaces of tables (the green box in \fig~\ref{fig:nyuv2}).} HF-cubes, LRRFast, and PCV can well smooth noisy surfaces but fail to preserve tiny details (the dashed circle in \fig~\ref{fig:nyuv2}).
Their long runtime also makes them less practical. 
Unsupervised NeuralGF is sensitive to noise. This will be further demonstrated in the subsequent experiments with synthetic data (Section \ref{sec:PCPdata}).
PointNorm-Net, on the other hand, copes with all these challenges. 
This experiment further illustrates that our method has better generalization performance. 

\subsubsection{Evaluation on sparse LiDAR dataset}
We further demonstrate our performance on the sparse LiDAR dataset: KITTI. Since 2012, it has been widely used as a public benchmark for various autonomous driving applications. 
In particular, the dataset provides sequentially organized point clouds collected by LiDAR with 64 beams. The experiments are conducted on the KITTI calibration sequence 2011-09-30. We use the device poses to splice out the whole scene, where the point cloud normals of every single frame are calculated and merged for the visualization in \fig~\ref{fig:teaser}.
Due to the significant time consumption of NeuralGF, it would take approximately 9 days to run the entire sequence. Therefore, only the point cloud frames 0-49 are evaluated for the visual comparison in \fig~\ref{fig:KITTI-1}.

This database does not have ground truth normals. For a more precise comparison, we cut out a portion of the structure composed of planes, such as a box\jz{,} and reconstruct the underlying surface to obtain the ground truth normals manually. Specifically, we first obtain the points on each plane manually and fit the plane by using the least squares method. Then, the fitted plane is fine-tuned with the help of MeshLab. Finally, all the points on each plane are aggregated together, and the normal of the plane closest to each point is regarded as the true normal of that point. \fig~\ref{fig:KITTI-1} shows visual comparisons of estimated normals rendered in RGB color and the RMS is displayed below each result. More visual comparisons are presented in the supplementary material.

\begin{figure*}[htbp]
\vspace{-0.2cm} 
\setlength{\abovecaptionskip}{0cm}
\begin{center}
\begin{tabular}{@{}c@{} }
\includegraphics[width=1\textwidth, keepaspectratio]{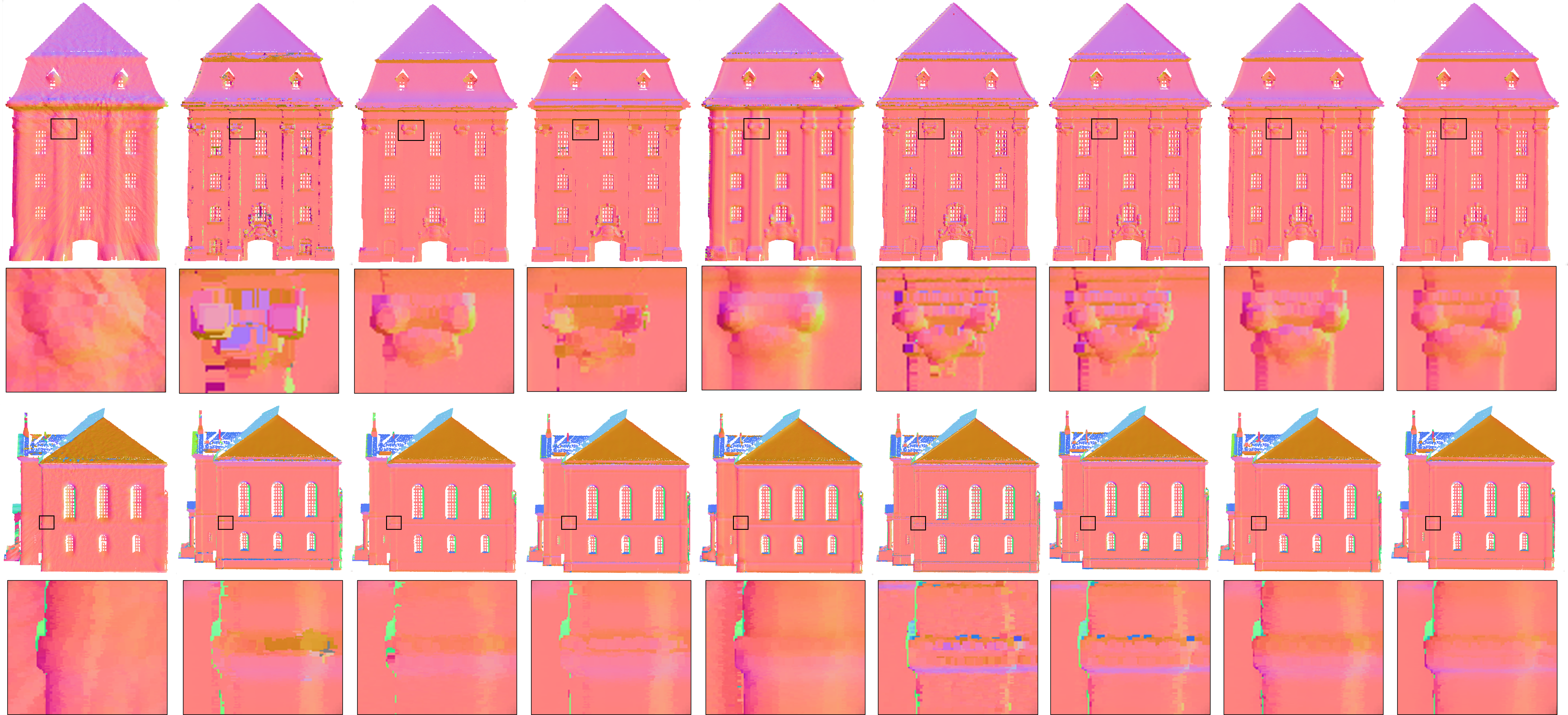}
\put(-510,-10){NeuralGF}
\put(-449,-10){LRRFast}
\put(-395,-10){HF-cubes}
\put(-326,-10){PCV}
\put(-275,-10){PCPNet}
\put(-215,-10){AdaFit}
\put(-160,-10){DeepFit}
\put(-112,-10){PointNorm}
\put(-97,-20){-Opt}
\put(-55,-10){PointNorm}
\put(-40,-20){-Net}
\end{tabular}
\end{center}
\caption{Visual comparison on the Semantic3D dataset. The point normal vectors are mapped to RGB colors.} 

\label{fig:Semantic3D}
\vspace{0.1cm} 
\end{figure*}

\begin{table*}[htbp]\small
\begin{center}
\caption{Comparison of the RMS angle error on the synthetic dataset of PCPNet. The best and second best results are shown in red and blue fonts respectively.}
\vspace{-0.2cm} 
\label{tab:RMS_PCP}
\begin{tabular}{m{0.13\textwidth}<{\raggedright}|m{0.08\textwidth}<{\centering } m{0.08\textwidth}<{\centering } m{0.08\textwidth}<{\centering } | m{0.08\textwidth}<{\centering } m{0.08\textwidth}<{\centering } m{0.08\textwidth}<{\centering }|m{0.08\textwidth}<{\centering } | m{0.08\textwidth}<{\centering }}
\hline
  &  \multicolumn{3}{c|}{\textbf{No noise}} &\multicolumn{3}{c|}{\textbf{Noise}} & & \textbf{Time}  \\ 
\cline{1-8}  
Method       & Uniform & Gradient & Stripes   & Small & Middle & Large         & Average  & (ms/point) \\ 
\hline
HF-cubes       &14.42   &14.84 &16.1 &13.68 &18.86 &27.68 &17.59 & \textcolor{blue}{\textbf{0.49}} \\
LRRfast         &11.57 &12.55 &13.4 & 13.09 &19.33 &26.75&16.11 &5.97\\
PCV            &13.54 & 14.27& 14.52 &14.51 &\textcolor{blue}{\textbf{18.5}} & 26.65 & 16.99 &2.36 \\
NeuralGF       &\textcolor{red}{\textbf{7.89}}  &\textcolor{red}{\textbf{9.21}}  &\textcolor{red}{\textbf{9.29}}  & \textcolor{red}{\textbf{9.85}}  &18.62 &\textcolor{blue}{\textbf{24.89}} & \textcolor{red}{\textbf{13.29}} &  12.57\\
PointNorm-Opt  &9.8 &10.47&10.91  & 13.07 &19.37& 25.55&14.86 &103.26\\
PointNorm-Net & \textcolor{blue}{\textbf{8.69}} & \textcolor{blue}{\textbf{9.62}} &\textcolor{blue}{\textbf{10.33}} 
 &\textcolor{blue}{\textbf{11.32}}  & \textcolor{red}{\textbf{17.58}} &\textcolor{red}{\textbf{23.93}}   & \textcolor{blue}{\textbf{13.57}} & \textcolor{red}{\textbf{0.36}} \\
\hline
\end{tabular}
\end{center}
\vspace{-0.6cm} 
\end{table*}

The LiDAR point clouds show significant noise unobserved in synthetic data. NeuralGF performs poorly, with high RMS errors. 
\re{Compared with DeepFit, a classical supervised method, and PCA, the most widely used technique, both SHS-Net and CMG-Net demonstrate enhanced capability in handling the KITTI dataset which exhibits sparsity and high noise levels. PointNorm-Net achieves the lowest RMS value while demonstrating superior visualization quality.}

\subsubsection{Evaluation on dense TLS dataset}

The Semantic3D \cite{Semantic3D} dataset provides 30 non-overlapping outdoor scenes acquired with a Terrestrial Laser Scanner (TLS).  
We mainly report visual comparisons on this dataset, see \fig~\ref{fig:Semantic3D}, since there are no ground truth normals. 
Compared to point clouds acquired via structure-from-motion pipelines, Kinect-like structured light sensors, and mobile LiDAR scanners, stationary TLS deliver significantly superior data quality in terms of density and accuracy \cite{Semantic3D}.
There is no observable domain gap between SceneNN and the synthetic training data.
Supervised DeepFit and AdaFit outperform PointNorm-Net in capturing details. However, PointNorm-Net still stands out as the leading option among unsupervised and self-supervised methods. It even outperforms the supervised PCPNet.

\begin{figure*}[htbp]
\setlength{\abovecaptionskip}{0.9cm}
\centering\footnotesize
\begin{overpic}[width=1\textwidth, keepaspectratio]{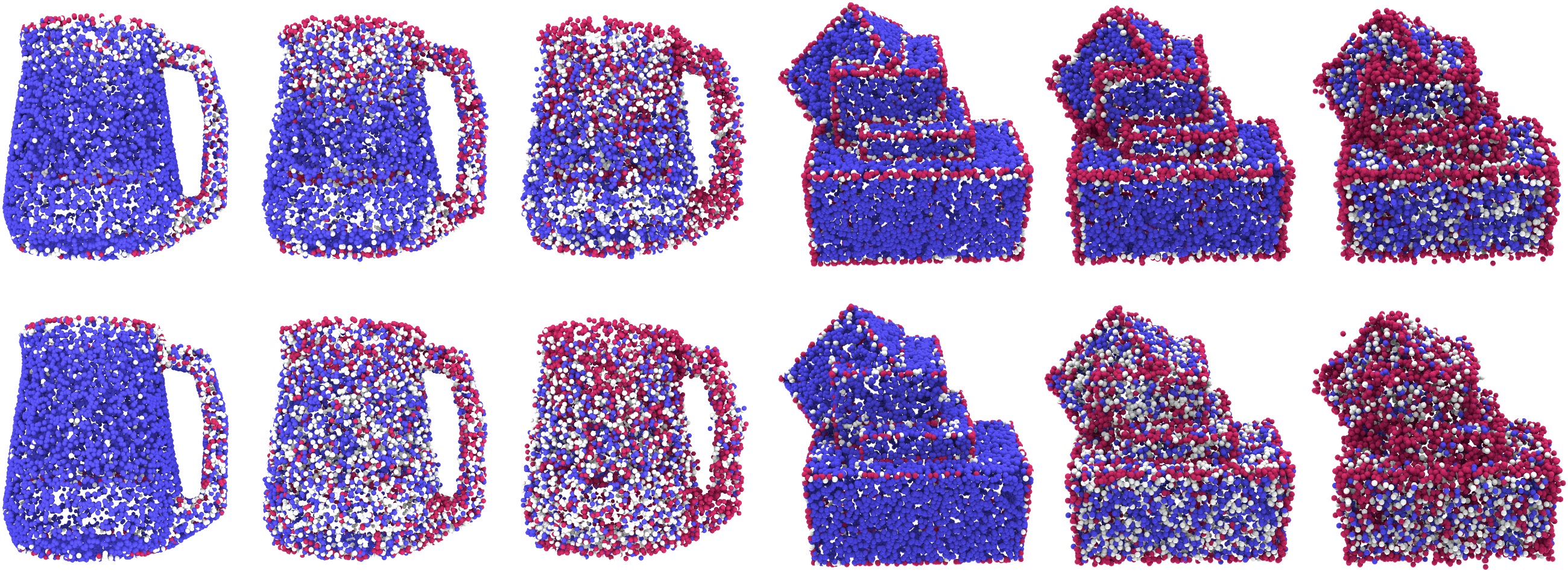}

\put(2,18){\small \textbf{0.8101/0.9460}}
\put(19,18){\small 0.6272/0.8461}
\put(35,18){\small 0.4033/0.6926}
\put(53,18){\small \textbf{0.7320/0.8234}}
\put(70,18){\small 0.5619/0.6703}
\put(87,18){\small 0.3335/0.5556}

\put(2,-1){\small 0.8620/0.9565}
\put(19,-1){\small \textbf{0.4025/0.7620}}
\put(35,-1){\small \textbf{0.2162/0.5427}}
\put(53,-1){\small 0.8563/0.9168}
\put(70,-1){\small \textbf{0.3282/0.6634}}
\put(87,-1){\small \textbf{0.1494/0.4042}}

\put(3,-3){\normalsize Low-noise}
\put(18.5,-3){\normalsize Medium-noise}
\put(35.8,-3){\normalsize High-noise}
\put(54,-3){\normalsize Low-noise}
\put(69.5,-3){\normalsize Medium-noise}
\put(88,-3){\normalsize High-noise}

\end{overpic}
    \caption{Visual comparison of the estimated normals of unsupervised NeuralGF (bottom row) and self-supervised PointNorm-Net (top row). The shapes are from the PCPNet dataset and colored according to the normal angle error, where darker reds indicate larger errors and darker blues indicate smaller errors. {The PGP5/PGP10 metrics are shown below each result.}}
\vspace{-0.2cm} 
\label{fig:PGP-5-10}
\end{figure*}

\begin{table*}[htbp]\small
\begin{center}
\vspace{0.3cm} 
\caption{Comparison of RMS angle error on point clouds sampled from triangular meshes. The best and second best results in unsupervised and self-supervised methods are shown in red and blue fonts respectively. The best results in supervised methods are bold.}
\vspace{-0.1cm} 
\label{tab:RMS_SceneNN}
\begin{tabular}{m{0.14\textwidth}<{\raggedright}|m{0.05\textwidth}<{\centering } m{0.05\textwidth}<{\centering }  m{0.05\textwidth}<{\centering } | m{0.06\textwidth}<{\centering } m{0.04\textwidth}<{\centering } m{0.075\textwidth}<{\centering } m{0.05\textwidth}<{\centering } m{0.125\textwidth}<{\centering } m{0.125\textwidth}<{\centering }}
\hline
  &  \multicolumn{3}{c|}{\textbf{Supervised}} &\multicolumn{5}{c}{\textbf{Unsupervised and self-supervised}} \\ 
\hline  
Method      & PCPNet & DeepFit   &  AdaFit  & NeuralGF & PCV & HF-cubes   & LRRfast  & PointNorm-Opt & PointNorm-Net \\ 
\hline
SceneNN          & 17.87 & 10.33   &  \textbf{8.39}  & 30.67   &20.28 &  18.59    & 18.79  & \textcolor{blue}{\textbf{11.72}}  & \textcolor{red}{\textbf{11.70}} \\
PCPNet's dataset & 14.34 & 11.8   &  \textbf{10.76}  & \textcolor{red}{\textbf{13.29}}   &16.99 &  17.59    & 16.11 &14.86   &  \textcolor{blue}{\textbf{13.57}} \\
\hline
\end{tabular}
\end{center}
\end{table*}

\subsection{Evaluation on synthetic point clouds sampled from triangular meshes}\label{sec:input}
\subsubsection{Evaluation on the PCPNet dataset}\label{sec:PCPdata}
The PCPNet dataset \cite{PCPNet:2018} is composed of point clouds densely sampled from triangular meshes, 100k points per shape. The point clouds are augmented by introducing Gaussian noise for each point’s spacial location with a standard deviation of 0.012, 0.006, 0.00125 w.r.t the bounding box. In the test set, two additional point clouds for each mesh are sampled with varying densities.
RMS angle error of our methods (PointNorm-Opt and PointNorm-Net) and other unsupervised methods on the PCPNet test set are shown in \tab~\ref{tab:RMS_PCP}. Both PointNorm-Opt and PointNorm-Net estimate more accurate normals than all conventional methods. 
Although NeuralGF outperforms our methods on no-noise and low-noise data, it is sensitive to medium-noise and high-noise data which is common in real data. Some visualization can be found in  \fig~\ref{fig:PGP-5-10}. In addition, LRRfast, NeuralGF, and PointNorm-Opt are very costly.

The comparison with supervised learning is shown in \tab~\ref{tab:RMS_SceneNN}.  It can be seen that supervised methods based on deep learning generally perform better than traditional unsupervised methods because they use ground truth normals for training. Our self-supervised method based on deep learning is better than supervised PCPNet, and less effective than supervised AdaFit and DeepFit.
However, they show degraded performance on real-world point clouds due to the domain gap between synthetic point clouds sampled from meshes and actual sensor measurements.

\begin{figure*}[htbp]
\setlength{\abovecaptionskip}{0.0cm}
\begin{center}
\begin{tabular}{@{}c@{} }
\includegraphics[width=1\textwidth, keepaspectratio]{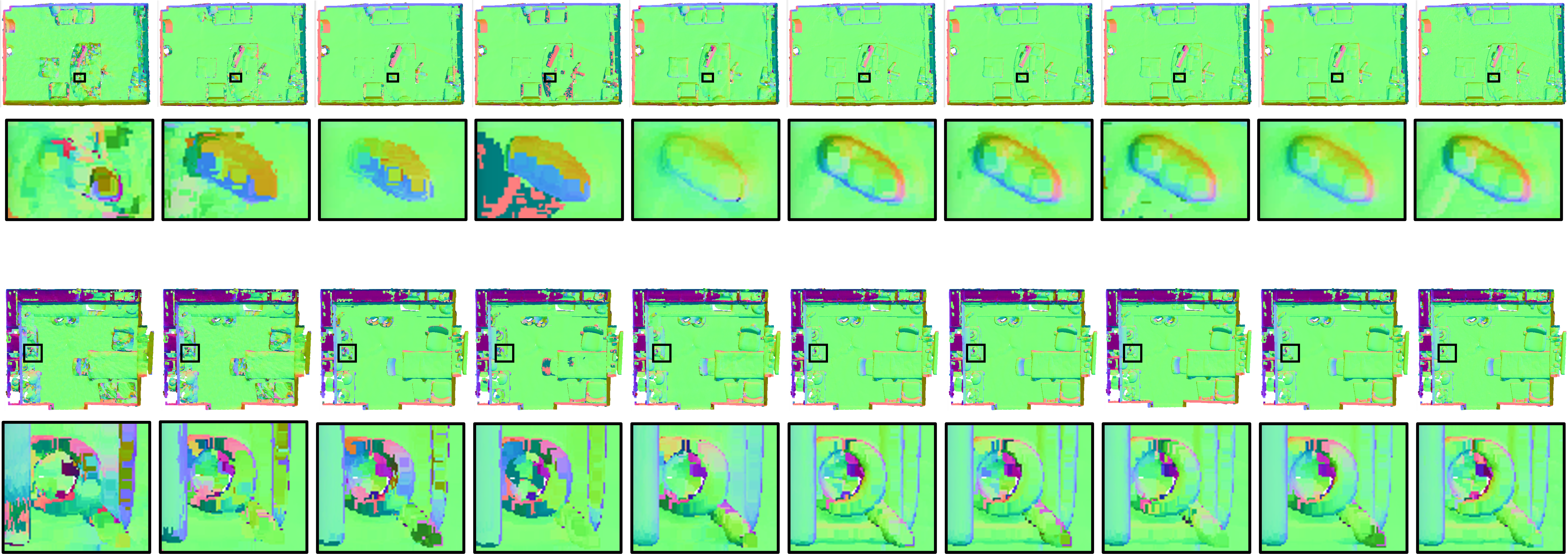}
\put(-513,101){NeuralGF}
\put(-458,101){LRRFast}
\put(-410,101){HF-cubes}
\put(-347,101){PCV}
\put(-302,101){PCPNet}
\put(-248,101){AdaFit}
\put(-198,101){DeepFit}
\put(-154,101){PointNorm}
\put(-140,91){-Opt}
\put(-102,101){PointNorm}
\put(-87,91){-Net}
\put(-43,101){Ground}
\put(-38,91){truth}

\put(-513,-10){NeuralGF}
\put(-458,-10){LRRFast}
\put(-410,-10){HF-cubes}
\put(-347,-10){PCV}
\put(-302,-10){PCPNet}
\put(-248,-10){AdaFit}
\put(-198,-10){DeepFit}
\put(-154,-10){PointNorm}
\put(-140,-20){-Opt}
\put(-102,-10){PointNorm}
\put(-87,-20){-Net}
\put(-43,-10){Ground}
\put(-38,-20){truth}
\end{tabular}
\end{center}
\caption{Visual comparison on the SceneNN dataset. The point
normal vectors are mapped to RGB colors. 
    }
\label{fig:SceneNN}
\vspace{-0.2cm} 
\end{figure*}

\subsubsection{Evaluation on the SceneNN dataset} 
The SceneNN dataset \cite{SceneNN} consists of indoor scenes captured by a depth camera. 
We use the point clouds provided by HSurf-Net \cite{lihsurf}, which are sampled from the reconstructed meshes, and use the mesh normals as the ground-truth normals.  
In the unsupervised and self-supervised methods, our methods obtain the lowest RMS, see \tab~\ref{tab:RMS_SceneNN}, and more robust at regions with sharp features, see \fig~\ref{fig:SceneNN}. 
They also outperform the supervised PCPNet. Compared to the supervised DeepFit and AdaFit, they are less effective. This is mainly because the point clouds are sampled from meshes and share similar characteristics with the synthetic training data from PCPNet.

\begin{table}[htbp]\small
\begin{center}
\caption{Ablation study of PointNorm-Net on the dataset of PCPNet.}
\label{tab:RMS_ablation_loss}
\vspace{-0.2cm} 
\begin{tabular}{c|c|c|c|c|c|c}
\hline
Row number&  CNI & ANS & FCF  &$L_{ccn}$ &         $ L_{w}$  &  RMS$\downarrow$\\
\hline
1 &                  &                 &             &           &                   &        17.00     \\
\hline
2 &     \checkmark   &                &             &            &                   &      16.46   \\
\hline
3 & \checkmark       &    \checkmark &               &           &                   &    14.79      \\
 \hline
4 &\checkmark    &    \checkmark &   \checkmark     &                &                   &    14.28  \\
\hline
5 &   \checkmark   &    \checkmark  &   \checkmark      & \checkmark      &                   &      13.97  \\
\hline
6 &   \checkmark   &    \checkmark  &   \checkmark      &               &    \checkmark  &      13.73 \\
\hline
7 &   \checkmark   &    \checkmark  &   \checkmark      &  \checkmark &    \checkmark  &      13.57 \\
\hline
\end{tabular}
\end{center}
\vspace{-0.3cm} 
\end{table}

\begin{table*}[htbp]\small
\begin{center}
\vspace{0.2cm} 
\caption{Comparison of the RMS angle error for different rejection ratios on the synthetic dataset of PCPNet.}
\vspace{-0.2cm} 
\label{tab:proportion_rejection}
\begin{tabular}{c|c|c|c|c|c|c|c}
\hline
 & \multicolumn{3}{c|}{\textbf{No noise}} & \multicolumn{3}{c|}{\textbf{Noise}} &   \\ 
\hline
  \textbf{Rejection} &Uniform & Gradient& Stripes& Small & Middle & Large & Average\\ 
\hline
 0$\%$ & 9.33&10.15&	10.79& 11.69& 17.63	& 24.59	&14.03\\
\hline
 5$\%$ &9.31&	10.18&	10.75&11.77&	17.62&	24.58&	14.03\\
\hline
10$\%$ & 8.69  & 9.62 &10.33& 11.32  & 17.58 &\textbf{23.93} & \textbf{13.57} \\
\hline
20$\%$ &\textbf{8.59}&	\textbf{9.58}	&\textbf{10.14}	&	11.27	&\textbf{17.57}	&24.47&13.60\\
\hline
50$\%$ &8.63&9.56	&10.29	&\textbf{11.14}	&\textbf{17.57}&	24.47		&13.61 \\
\hline
\end{tabular}
\end{center}
\end{table*}
\begin{table*}[htbp]\small  
\begin{center}
\caption{Comparison of the RMS angle error for different number of segments on the synthetic dataset of PCPNet.}
\vspace{-0.2cm} 
\label{tab:neigh_seg}
\begin{tabular}{c|c|c|c|c|c|c|c}
\hline
 & \multicolumn{3}{c|}{\textbf{No noise}} & \multicolumn{3}{c|}{\textbf{Noise}} &   \\ 
\hline
 \textbf{Parameters setting} &Uniform & Gradient & Stripes& Small & Middle & Large& Average\\ 
\hline
Segments number = 2 & \multirow{3}{*}{8.8}&\multirow{3}{*}{9.71}&\multirow{3}{*}{10.28}& \multirow{3}{*}{11.36}	&\multirow{3}{*}{17.4}&	\multirow{3}{*}{25.32}		&\multirow{3}{*}{13.81 }\\
$k_{1}=32$,$k_{2}=128$, $k_{3} = -$ $k_{4}=-$ & & & & & & & \\
$l_0=0$, $l_1=0.02$, $l_2=0.3$, $l_3=-$, $l_4=-$ & & & & & & & \\
\hline
Segments number = 3 &\multirow{3}{*}{8.66} &\multirow{3}{*}{9.59}&\multirow{3}{*}{10.23}&\multirow{3}{*}{11.14 }& \multirow{3}{*}{17.17}	& \multirow{3}{*}{23.99}		&\multirow{3}{*}{13.46}\\
 $k_{1}=32$,$k_{2}=128$, $k_{3} =256$, $k_{4}=-$& & & & & & & \\
$l_0=0$, $l_1=0.02$, $l_2=0.14$, $l_3=0.3$, $l_4=-$ & & & & & & & \\
\hline
Segments number = 4 & \multirow{3}{*}{8.69}& \multirow{3}{*}{9.62} &\multirow{3}{*}{10.33} & \multirow{3}{*}{11.32}  & \multirow{3}{*}{17.58} &\multirow{3}{*}{23.93}  & \multirow{3}{*}{13.57} \\
 $k_{1}=32$,$k_{2}=128$, $k_{3} =256$,$k_{4}=450$ & & & & & & & \\
$l_0=0$, $l_1=0.02$, $l_2=0.14$, $l_3=0.16$, $l_4=0.3$ & & & & & & & \\ 
\hline
\end{tabular}
\end{center}
\end{table*}

\subsection{Efficiency}\label{sec:Efficiency}
In \fig~\ref{fig:teaser}, we report the average inference time of the different methods.
The deep learning methods are implemented in Python 3.7.3, while PCV and LRRfast are implemented in MATLAB, and HF-cubes is implemented in C++. 
\re{SHS-Net and MSECNet, the latest high-performing normal estimators, achieve the fastest inference speeds but exhibit significant performance degradation on real-world scanned datasets. 
DeepFit, G2N2, and PointNorm-Net have similar networks and inference times, taking about 0.3ms per point.
CMG-Net, PCPNet, and AdaFit, due to their more complex network structures or increased number of parameters, exhibit slower inference times of 0.4ms, 0.61ms, and 0.71ms per point, respectively.}
NeuralGF is very time-consuming since it fits a global neural field from each single point cloud.
Traditional high-quality methods exhibit better generalization ability than deep supervised methods but are usually time-consuming.
PointNorm-Net, a deep network-based self-supervised method, integrates the merits of the two approaches. 

\subsection{Ablation study}
\label{sec:ablation}

\textbf{Validation of construction factors for candidate normals.} Three factors that may affect the quality of candidate normals are examined: candidate normal initialization (CNI), adaptive neighborhood size (ANS), and feasible candidate filtering (FCF).
We ablate them to explore their influence on normal estimation, {see rows 1 to 4 of \tab~\ref{tab:RMS_ablation_loss}}. 
Without CNI, a single candidate normal, fitting all the neighbor points by PCA, is used to train the network. When Adaptive Neighborhood Selection (ANS) is not employed, we fix the neighborhood size to 256. To isolate and analyze the effect of candidate normals, we replace the candidate consensus and feature-preserving losses with a general loss function $L_{pre}$ that quantifies the discrepancy between the predicted normal and candidate normals at each query point:
\begin{equation}\label{eq:normal_loss_pre}
L_{pre}=\sum_{\theta}^{}\|\hat{\mathbf{n}}_{t} \times \mathbf{n}^{t}_{\theta}\|_{F}^{2},
\end{equation}
where $\mathbf{n}^{t}_{\theta}$ represents the candidate normals which can be a single normal (without CNI), initial candidates (without FCF), or feasible candidates (with FCF).

Row 1 of \tab~\ref{tab:RMS_ablation_loss} shows the results of training a network from PCA with 256 neighbors. After adding CNI, the RMS angle error is 0.54 lower, which illustrates that learning from multi-candidate normals is superior to that from a single normal.
Among these three components, ANS contributes the most significant improvement, reducing the RMS angle error by 1.67 and down to 14.79. 
FCF further improves the quality of candidate normals by filtering out the blurry normals in the initial candidate normals, which reduces the RMS angle error by 0.51.

\textbf{Validation of loss terms. }
The contribution of $L_{ccn}$ and $L_{w}$ is illustrated in rows 5 to 7 of \tab~\ref{tab:RMS_ablation_loss}. 
When $L_{ccn}$ is disabled, $L_{pre}$ defined in \eq~\ref{eq:normal_loss_pre} is used to make sure that the network learns from the candidate normals. It is found that both $L_{ccn}$ and $L_{w}$ further boost the performance of PointNorm-Net.

\begin{table*}[htbp]\small
\begin{center}
\caption{Evaluation of optimization and learning strategies within our multi-modal distribution estimation paradigm, using the PCPNet synthetic dataset.
}
\vspace{-0.2cm} 
\label{tab:optimization_network_learning}
\begin{tabular}{m{0.15\textwidth}<{\centering}|m{0.2\textwidth}<{\centering } | m{0.5\textwidth}<{\centering } |  m{0.05\textwidth}<{\centering }}
\hline
                &    Method   &   Description  & RMS \\
\hline
\multirow{2}{*}{Optimizing directly}     &   PointNorm-Opt &  It estimates the normal of point by directly minimizing \eq~\ref{eq:8}	&  20.18\\ 	
\cline{2-4}
 & PointNorm-Opt-weight&	We optimize per point weight by directly minimizing the
candidate consensus loss add the weight regularization loss $L_{regw}$. & 	14.86 \\
\hline
\multirow{2}{3cm}{Point normal learning network} & PointNorm-PCPNet & The deep normal predictor follows the architecture of single scale PCPNet. & 14.82 \\
\cline{2-4}
 & PointNorm-Multi-PCPNet & The deep normal predictor follows the architecture of multi-scale PCPNet. & 14,36  \\
\hline
\multirow{2}{3cm}{Point weight learning network} & PointNorm-DeepFit
(PointNorm-Net)  & The deep normal predictor follows the architecture of DeepFit. & 13.57 \\
\cline{2-4}
 & PointNorm-AdaFit & The deep normal predictor follows the architecture of AdaFit. & 12.36 \\
\hline
\end{tabular}
\vspace{-0.2cm} 
\end{center}
\end{table*}

\begin{table*}[htbp]\small
\begin{center}
\vspace{0.3cm} 
\caption{The generalization ability of DeepFit and PointNorm-Net. The RMS angle errors show that PointNorm-Net outperforms DeepFit on unseen data, i.e. when the training and test datasets have different characteristics and noise distributions.}
\vspace{-0.2cm} 
\label{tab:RMS_generalization}
\begin{tabular}{m{0.15\textwidth}<{\raggedright}|m{0.15\textwidth}<{\centering } m{0.15\textwidth}<{\centering } |  m{0.15\textwidth}<{\centering }  m{0.15\textwidth}<{\centering }}
\hline
Train and Test &  \multicolumn{2}{c|}{\textbf{Within-Domain}} & \multicolumn{2}{c}{\textbf{Cross-Domain}}  \\
\hline
Train set &  PCV‘s dataset & PCPNet’s dataset & PCV‘s dataset & PCPNet’s dataset \\ 
\hline  
Test set   &   PCV‘s dataset  &   PCPNet’s dataset   &   PCPNet’s dataset   &   PCV‘s dataset
 \\ 
\hline
DeepFit   &   \textbf{9.8}   &    \textbf{11.8}         &     15.99  &  13.68      \\
PointNorm-Net   &   10.82      &  13.57           &      \textbf{15.53}&   \textbf{10.89} \\
\hline
\end{tabular}
\end{center}
\end{table*}

\textbf{Validation of the proportion of candidates rejection.} In Section \ref{sec:Sel_candi_normasls}, we employ the idea of kernel density estimation to perform preliminary rejection of the initial candidate normals. We observe that a larger rejection ratio does not necessarily yield better results during the filtering process as illustrated in \tab~\ref{tab:proportion_rejection}.
Setting an overly aggressive rejection ratio reduces the pool of initial candidate normals, which compromises the accuracy benefits gained from multi-sample consensus estimation.
There exists an optimal range for the rejection ratio. We chose to reject 10\% of the initial candidate normals. 

\textbf{Validation of the adaptive neighborhood partition.} In Section \ref{sec:Optimization_NeighborSize}, the proposed algorithm adaptively determines the neighborhood size according to the noise scale of the point cloud.
In our experimental setup, we partition the local neighborhood into four distinct scales. Our ablation studies demonstrate that varying the number of partitions between 2 and 4 yields statistically insignificant differences in performance, as shown in \tab~\ref{tab:neigh_seg}. Consequently, the 4-segments schedule is adopted based on the characteristics of the PCPNet dataset. Note that the model trained only on the PCPNet dataset with this option also performs well on the other datasets as reported in the experiments above.


\section{Discussion}
\subsection{Optimization vs. network learning.}
The proposed multi-modal normal distribution estimation paradigm can be integrated into deep learning or optimization-based normal estimation frameworks. When applied to deep frameworks, various patch-based networks can be used as the deep normal predictor in \fig~\ref{fig:flow}. 
These extended versions and their results are shown in \tab~\ref{tab:optimization_network_learning}.

\textbf{(a) Using a network outperforms optimizing directly.} Compared to PointNorm-Opt, PointNorm-Net incorporates a neural network component and auxiliary losses to achieve superior estimation accuracy. \tab~\ref{tab:optimization_network_learning} shows that using the point normal learning networks of PCPNet also improves the accuracy. The average RMS drops from PointNorm-opt's 14.86 to PointNorm-PCPNet's 14.82 and PointNorm-Multi-PCPNet's 14.36 respectively.  \\
\textbf{(b) Point normal learning vs Point weight learning.} 
Supervised normal learning approaches have shown that point weight learning methods consistently outperform point normal learning methods, and this also holds true in our case. PointNorm-Net, which follows the point weight learning approach, outperforms PointNorm-Multi-PCPNet, a point normal learning method. Their respective RMS errors are 13.57 and 14.36, as reported in \tab~\ref{tab:optimization_network_learning}.
But when we try to optimize per point weight without a network, denoted as PointNorm-opt-weight, the RMS increased from 14.86 to 20.18 as shown in \tab~\ref{tab:optimization_network_learning}. \\
\textbf{(c) More powerful networks lead to better accuracy.} From \tab~\ref{tab:optimization_network_learning}, it is clear that PointNorm-Multi-PCPNet outperforms PointNorm-PCPNet, and PointNorm-Adafit outperforms PointNorm-Net, i.e. PointNorm-Deepfit. DeepFit is a widely used network in normal estimation. It has a small scale, fast speed, and high performance. Therefore, we choose to use this network. But, there are still many networks that may perform better in terms of performance, such as Adafit\cite{Adafit}, or speed, such as DeepIterative\cite{DeepIterative}.   \\

\begin{table*}[htbp]\small
\begin{center}
\caption{The denoising performance of DMR is gradually improved by applying our strategies.}
\vspace{-0.2cm} 
\label{tab:Denoising}
\begin{tabular}{c|c|c|c|c|c|c|c|c}
\hline
 Noise  scale  & \multicolumn{2}{c|}{1$\%$} &\multicolumn{2}{c|}{2$\%$} & \multicolumn{2}{c|}{2.5$\%$} & \multicolumn{2}{c}{3$\%$}   \\
\hline
   Error metric  &  CD   &  P2S    &      CD  &  P2S   &  CD   &   P2S  & CD  & P2S  \\
\hline
DMR  &2.47 &2.97& 4.01& 5.58 & 4.92 &7.20 & 5.77 & 8.73 \\
\hline 
DMR+CPI  &2.38 & 2.84 & 3.74 & 5.15 & 4.61 & 6.65 & 5.49 & 8.24 \\
\hline
DMR+CPI+$L_{ccp}$ & 2.13 &2.43 &3.54 &4.75&4.42&6.25&5.27&7.78\\
\hline
DMR+CPI+$L_{ccp}$+FCF &\textbf{2.13} &\textbf{2.41} &\textbf{3.47} &\textbf{4.60} &\textbf{4.28} &\textbf{6.00}& \textbf{5.08} &\textbf{7.42}\\
\hline 
\end{tabular}
\end{center}
\vspace{-0.2cm} 
\end{table*}
\begin{figure}
\setlength{\abovecaptionskip}{0.4cm}
\centering\small
\begin{overpic}[width=0.5\textwidth, keepaspectratio]{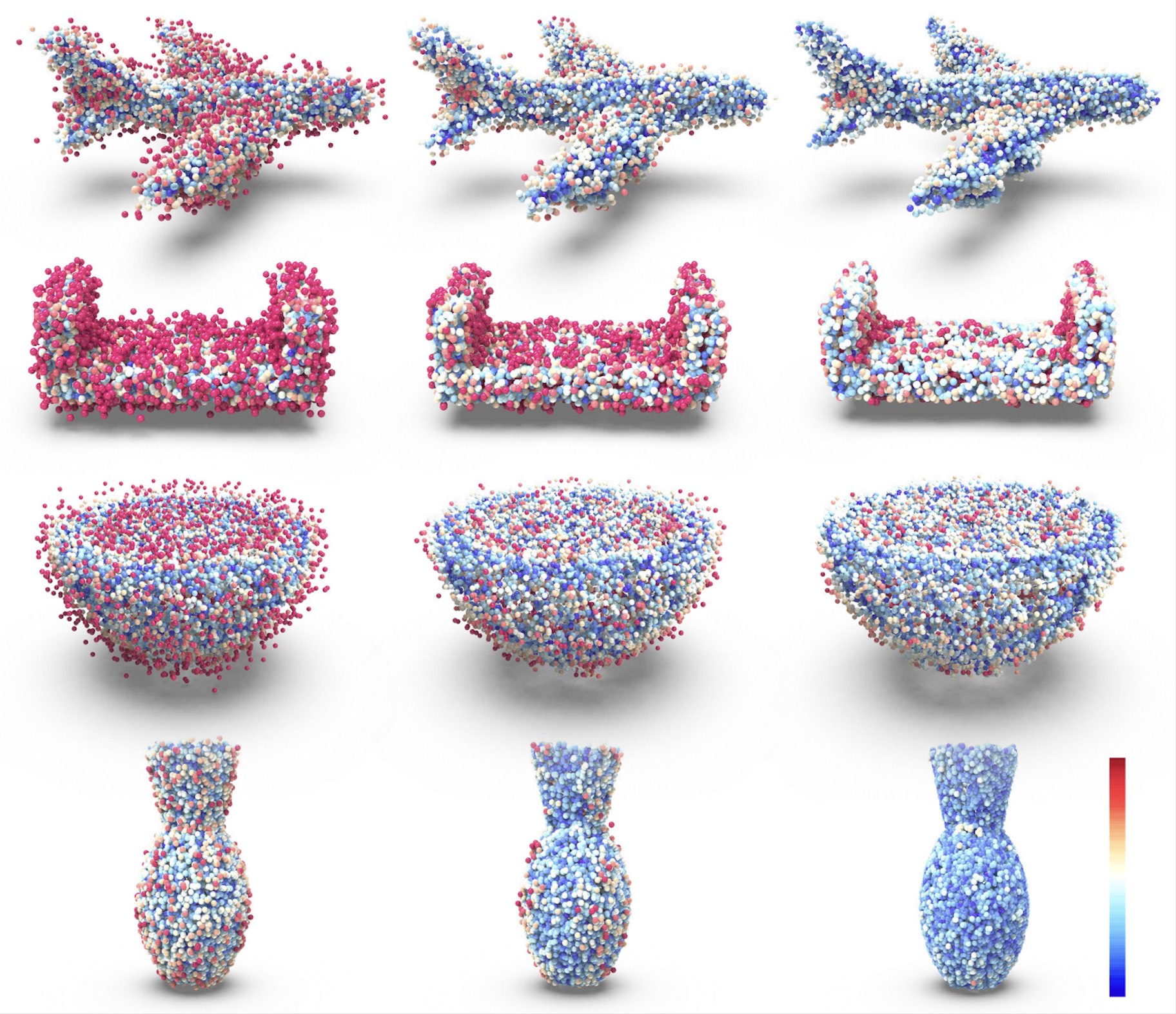}
\put(4,-2.5){Raw point cloud}
\put(44.8,-2.5){DMR}
\put(79.7,-2.5){Our}

\put(96.5,0.6){0 \footnotesize}
\put(95.9,21.1){Up \footnotesize}
\end{overpic}
    \caption{More visually pleasing surfaces are generated by the proposed denoising method (DMR+CPI+$L_{ccp}$+FCF) than the baseline method: DMR. The colors, from blue to red, encode the P2S error. 
}
\vspace{-0.5cm} 
\label{fig:denoise}
\end{figure}

\subsection{Generalization to different data.} 
Notably, supervised methods outperform traditional approaches and our proposed techniques when tested on datasets such as SceneNN and Semantic3D that have similar characteristics to the training data. However, when there are significant differences between the characteristics and noise distributions of the training and testing data (i.e., in the case of cross-domain evaluation), supervised deep learning methods perform worse than traditional approaches and our methods. These findings emphasize the necessity of further investigating the generalizability of supervised deep normal estimators. In \tab~\ref{tab:RMS_generalization}, we conduct a comparative analysis of PointNorm-Net and DeepFit. When both methods are trained on the PCV dataset, DeepFit attains lower RMS angle errors than PointNorm-Net on the PCV dataset (9.8 vs 10.82). Conversely, PointNorm-Net surpasses DeepFit on the PCPNet dataset (15.53 vs 15.99).

\subsection{Generalization to self-supervised point cloud denoising. }
\label{sec:extension}
The proposed multi-sample consensus paradigm is also applicable to other self-supervised tasks. This is shown by improving the cutting-edge self-supervised point cloud denoising method: DMR \cite{luo2020differentiable}. 
DMR follows the denoising loss defined in \cite{totaldenoise}:
\begin{equation}\label{eq:denoise_loss}
L_U=\frac{1}{N}\sum_{t=1}^{N}\sum_{\mathbf{q}}\|f(\mathbf{p}_t)-\mathbf{q}\|,
\end{equation}
where $f(\cdot)$ represents the denoiser that maps a noisy point to a denoised point. Points $\{\mathbf{q}\}$ are sampled from the noisy
input point cloud $P$ according to a prior $P(\mathbf{q}|\mathbf{p}_t)$, which is empirically defined as $P(\mathbf{q}|\mathbf{p}_t)\propto exp(-\frac{\|\mathbf{q}-\mathbf{p}_t\|}{2\omega ^2})$.
However, it is not a suitable assumption for points around the sharp features. Although Hermosilla \etal~\cite{totaldenoise} employ RGB color annotation to overcome this problem, it only reduces it but does not eliminate it completely. Moreover, RGB color annotation is not always available.

Here, we first design a novel candidate position initialization (CPI) strategy to generate multiple and more accurate sampling points $\{\mathbf{q}\}$. 
A candidate is infeasible if there are fewer neighbors of $\mathbf{p}_t$ near it. Therefore, we then use the Feasible candidate filtering (FCF) to obtain feasible candidates position. Finally, a candidate consensus loss for position, similar with $L_{ccn}$, is defined as follows:
\begin{equation}
L_{ccp}=\sum_{\mathbf{q}}^{}-e^{-\frac{\left\|f(\mathbf{p}_{t}) - \mathbf{q}\right\|_{F}^{2}}{\tau^{2}}},
\end{equation}
where $\tau= \sigma$. 
It encourages the results to converge to the major mode supported by more inlier candidates.

We evaluate the effectiveness of our strategies quantitatively and qualitatively on the dataset generated from ModelNet-40 \cite{modelnet-40}.
For quantitative comparison, the Chamfer distance (CD) and point-to-surface distance (P2S) are used as evaluation metrics. As shown in \tab~\ref{tab:Denoising}, the CD and P2S are gradually improved by applying the three strategies. 
A qualitative comparison is also shown in \fig~\ref{fig:denoise}, where the colors encode the P2S errors.  Our results are much cleaner and exhibit more visually pleasing surfaces than those of DMR. More details about the strategy design and data generationt can be found in the supplementary material.

\section{Conclusion}
In this paper, we prove that the normal of each query point is the expectation of randomly sampled candidate normals. Building on this, a novel major mode identification paradigm for multi-modal distributions, leveraging multisample consensus techniques, is proposed to boost self-supervised normal estimation.  
It can handle cases not covered in the theory, i.e. the initial candidate normals satisfy a multi-modal distribution for the existence of sharp features. 
The paradigm consists of three stages: candidate normal initialization, feasible candidate filtering, and major mode estimation.
We present two implementations of the paradigm: PointNorm-Opt, a novel optimization-based baseline, and PointNorm-Net, the first self-supervised deep learning approach for surface normal estimation. 
PointNorm-Net shows better generalization ability and significantly outperforms previous optimization methods and SOTA supervised and unsupervised deep normal estimators on three real datasets captured by Kinect and LiDAR. The paradigm can also be integrated into other low-level point cloud processing missions. The cutting-edge method DMR is gradually improved by applying our strategies.

Our approach has several limitations. Notably, the current method does not explicitly address challenges posed by thin geometric structures or non-uniform point cloud sampling densities. 
In the future, it may be possible to utilize the density of points to improve the accuracy of these two cases. Furthermore, we intend to conduct in-depth research in low-level image and point cloud analysis via self-supervised deep learning driven by multi-sample consensus. It would also be interesting to explore multiple pseudo-labels for semantic-level learning tasks.
\IEEEdisplaynontitleabstractindextext

%
\IEEEpeerreviewmaketitle

\ifCLASSOPTIONcaptionsoff
  \newpage
\fi

\begin{IEEEbiography}[{\includegraphics[width=1in,height=1.25in,clip,keepaspectratio]{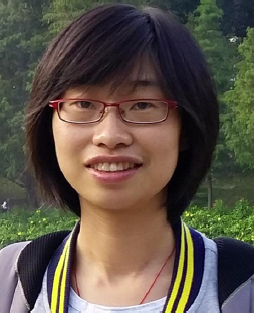}}]{Jie Zhang}
received the PhD degree in 2015 from the Dalian University of Technology, China.
She is currently an associate professor with the School of Mathematics, Liaoning Normal University, China. Her current research Interests
include geometric processing and machine learning.
%
\end{IEEEbiography}

\begin{IEEEbiography}[{\includegraphics[width=1in,height=1.25in,clip,keepaspectratio]{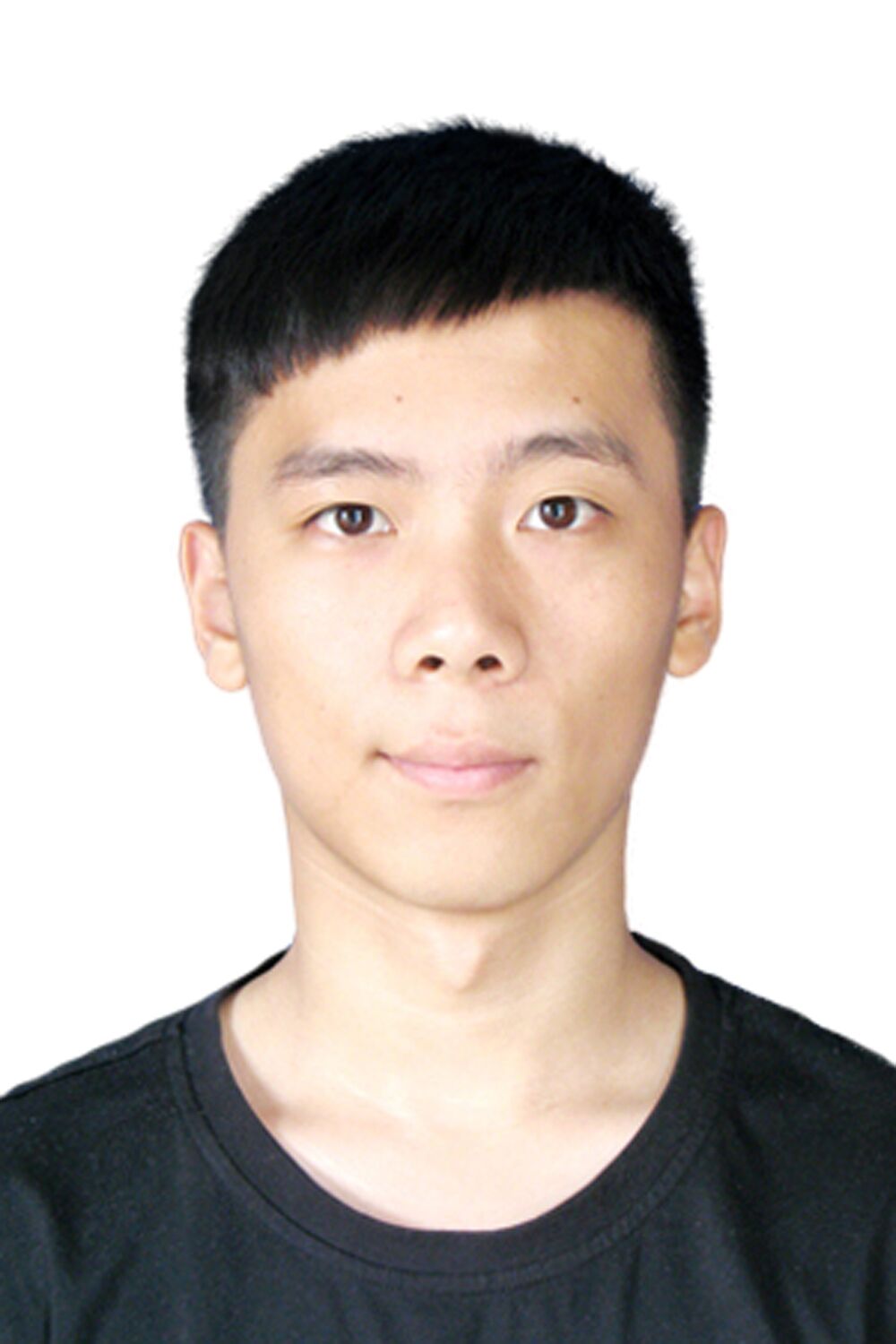}}]{Minghui Nie}
is currently a Ph.D. student in software engineering at Jiangnan University, China. He earned his master’s degree in computational mathematics from Liaoning Normal University, China, in 2024, and received the bachelor’s degree in mathematics and applied mathematics from Dezhou University, China, in 2021. His research interests include geometric processing and machine learning.
\end{IEEEbiography}

\begin{IEEEbiography}[{\includegraphics[width=1in,height=1.25in,clip,keepaspectratio]{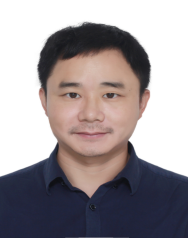}}]{Changqing Zou}
received the BE degree from the Harbin Institute of Technology, Harbin, China, the ME degree from the Institute of Remote Sensing and Digital Earth, Chinese Academy of Sciences, Beijing, China, and the PhD degree from the Shenzhen Institutes of Advanced Technology, Chinese Academy of Sciences, Beijing, China. He is currently a 
ZJU100 Young Professor at Zhejiang University. His research interests include computer graphics, computer vision, and machine learning.
\end{IEEEbiography}%

\begin{IEEEbiography}[{\includegraphics[width=1in,height=1.25in,clip,keepaspectratio]{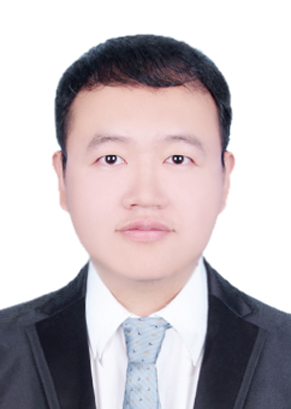}}]{Jian Liu}
received the PhD degree in 2022 from the Shandong University, China. He served as a postdoctoral researcher at the School of Software, Tsinghua University from 2022 to 2024. Currently, he is a lecturer at the School of Software, Shenyang University of Technology. His current research interests include geometric processing and machine learning.
\end{IEEEbiography}%

\begin{IEEEbiography}[{\includegraphics[width=1in,height=1.25in,clip,keepaspectratio]{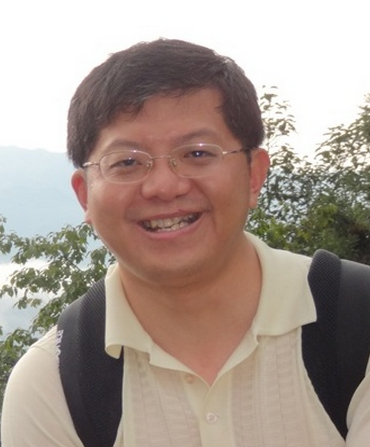}}]{Ligang Liu}
received the BSc degree in 1996 and the PhD degree in 2001 from Zhejiang University,
China. He is a professor at the University of Science and Technology, China. Between 2001 and 2004, he was at Microsoft Research Asia. Then he was at Zhejiang University during 2004 and 2012. He paid an academic visit to Harvard University during 2009 and 2011. His research interests include geometric processing and image processing. His research works can be found at his research website: http://staff.ustc.edu.cn/lgliu. 
\end{IEEEbiography}%

\begin{IEEEbiography}[{\includegraphics[width=1in,height=1.25in,clip,keepaspectratio]{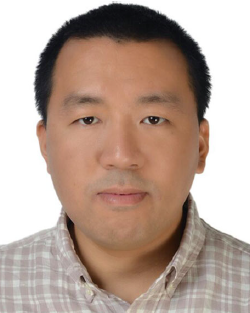}}]{Junjie Cao}
received the BSc degree in 2003 and the PhD degree in 2010 from Dalian University of Technology, China. He is an associate professor with the School of Mathematical Sciences, Dalian University of Technology, China. Between 2014 and 2015, he paid an academic visit to Simon Fraser University, Canada. His research interests include geometric processing and image processing. 
His research works can be found at his research website: https://jjcao.github.io/.
\end{IEEEbiography}%




\bibliographystyle{IEEEtran}
\bibliography{nrefs}

\begin{thebibliography}{10}
\providecommand{\url}[1]{#1}
\csname url@samestyle\endcsname
\providecommand{\newblock}{\relax}
\providecommand{\bibinfo}[2]{#2}
\providecommand{\BIBentrySTDinterwordspacing}{\spaceskip=0pt\relax}
\providecommand{\BIBentryALTinterwordstretchfactor}{4}
\providecommand{\BIBentryALTinterwordspacing}{\spaceskip=\fontdimen2\font plus
\BIBentryALTinterwordstretchfactor\fontdimen3\font minus \fontdimen4\font\relax}
\providecommand{\BIBforeignlanguage}[2]{{%
\expandafter\ifx\csname l@#1\endcsname\relax
\typeout{** WARNING: IEEEtran.bst: No hyphenation pattern has been}%
\typeout{** loaded for the language `#1'. Using the pattern for}%
\typeout{** the default language instead.}%
\else
\language=\csname l@#1\endcsname
\fi
#2}}
\providecommand{\BIBdecl}{\relax}
\BIBdecl

\bibitem{Rui2023GCNO}
R.~Xu, Z.~Dou, N.~Wang, S.~Xin, S.~Chen, M.~Jiang, X.~Guo, W.~Wang, and C.~Tu, ``Globally consistent normal orientation for point clouds by regularizing the winding-number field,'' \emph{ACM Trans. Graph.}, vol.~42, no.~4, pp. 1--15, 2023.

\bibitem{MultiNormal2019}
J.~{Zhang}, J.~{Cao}, X.~{Liu}, H.~{Chen}, B.~{Li}, and L.~{Liu}, ``Multi-normal estimation via pair consistency voting,'' \emph{IEEE Trans. Vis. Comput. Graph.}, vol.~25, no.~4, pp. 1693--1706, 2019.

\bibitem{HF-cubes}
A.~Boulch and R.~Marlet, ``Fast and robust normal estimation for point clouds with sharp features,'' \emph{Comput. Graph. Forum}, vol.~31, no.~5, pp. 1765--1774, 2012.

\bibitem{PCPNet:2018}
P.~Guerrero, Y.~Kleiman, M.~Ovsjanikov, and N.~J. Mitra, ``{PCPNet}: Learning local shape properties from raw point clouds,'' \emph{Comput. Graph. Forum}, vol.~37, no.~2, pp. 75--85, 2018.

\bibitem{zhou2022refine}
H.~Zhou, H.~Chen, Y.~Zhang, M.~Wei, H.~Xie, J.~Wang, T.~Lu, J.~Qin, and X.-P. Zhang, ``{Refine-Net}: Normal refinement neural network for noisy point clouds,'' \emph{IEEE Trans. Pattern Anal. Mach. Intell.}, vol.~45, pp. 946--963, 2022.

\bibitem{DeepFit}
Y.~Ben{-}Shabat and S.~Gould, ``{DeepFit}: 3d surface fitting via neural network weighted least squares,'' in \emph{Eur. Conf. Comput. Vis.}, vol. 12346, 2020, pp. 20--34.

\bibitem{Adafit}
R.~Zhu, Y.~Liu, Z.~Dong, T.~Jiang, Y.~Wang, W.~Wang, and B.~Yang, ``{AdaFit}: Rethinking learning-based normal estimation on point clouds,'' in \emph{{IEEE/CVF} Int. Conf. Comput. Vis}, 2021, pp. 6118--6127.

\bibitem{NestiNet2019}
Y.~Ben-Shabat, M.~Lindenbaum, and A.~Fischer, ``{Nesti-Net}: Normal estimation for unstructured 3d point clouds using convolutional neural networks,'' in \emph{{IEEE} Conf. Comput. Vis. Pattern Recogn.}, 2019, pp. 10\,104--10\,112.

\bibitem{Geometry-Guided}
J.~Zhang, J.-J. Cao, H.-R. Zhu, D.-M. Yan, and X.-P. Liu, ``Geometry guided deep surface normal estimation,'' \emph{Comput.-Aided Des.}, vol. 142, no. 103119, 2022.

\bibitem{DeepIterative}
J.~E. Lenssen, C.~Osendorfer, and J.~Masci, ``Deep iterative surface normal estimation,'' in \emph{{IEEE/CVF} Conf. Comput. Vis. Pattern Recogn.}, 2020, pp. 11\,244--11\,253.

\bibitem{lihsurf}
Q.~Li, Y.-S. Liu, J.-S. Cheng, C.~Wang, Y.~Fang, and Z.~Han, ``{HSurf-Net}: Normal estimation for 3d point clouds by learning hyper surfaces,'' in \emph{Adv. in Neural Inf. Process. Syst.}, 2022.

\bibitem{MDRNet}
J.~Cao, H.~Zhu, Y.~Bai, J.~Zhou, J.~Pan, and Z.~Su, ``Latent tangent space representation for normal estimation,'' \emph{IEEE Transactions on Industrial Electronics}, vol.~69, pp. 921 -- 929, 2022.

\bibitem{least_squares}
X.~Liu, J.~Zhang, J.~Cao, B.~Li, and L.~Liu, ``Quality point cloud normal estimation by guided least squares representation,'' \emph{Comput. Graph.}, vol.~51, pp. 106--116, 2015.

\bibitem{Zhang13}
J.~Zhang, J.~Cao, X.~Liu, J.~Wang, J.~Liu, and X.~Shi, ``Point cloud normal estimation via low-rank subspace clustering,'' \emph{Comput. Graph.}, vol.~37, no.~6, pp. 697--706, 2013.

\bibitem{Bao}
B.~Li, R.~Schnabel, R.~Klein, Z.~Cheng, G.~Dang, and S.~Jin, ``Robust normal estimation for point clouds with sharp features,'' \emph{Comput. Graph.}, vol.~34, no.~2, pp. 94--106, 2010.

\bibitem{Jets}
F.~Cazals and M.~Pouget, ``Estimating differential quantities using polynomial fitting of osculating jets,'' \emph{Comput. Aided Geom. Des.}, vol.~22, no.~2, pp. 121--146, 2005.

\bibitem{HoppeDDMS92}
H.~Hoppe, T.~DeRose, T.~Duchamp, J.~A. McDonald, and W.~Stuetzle, ``Surface reconstruction from unorganized points,'' in \emph{Annu. Conf. Comput. Graph. Interact. Tech.}, 1992, pp. 71--78.

\bibitem{Co-supported20}
H.~Zhou, H.~Chen, Y.~Feng, Q.~Wang, J.~Qin, H.~Xie, F.~L. Wang, M.~Wei, and J.~Wang, ``Geometry and learning co-supported normal estimation for unstructured point cloud,'' in \emph{{IEEE/CVF} Conf. Comput. Vis. Pattern Recogn.}, 2020, pp. 13\,235--13\,244.

\bibitem{geiger2012we}
A.~Geiger, P.~Lenz, and R.~Urtasun, ``Are we ready for autonomous driving? the kitti vision benchmark suite,'' in \emph{{IEEE} Conf. Comput. Vis. Pattern Recogn.}, 2012, pp. 3354--3361.

\bibitem{nyuv22012}
N.~Silberman, D.~Hoiem, P.~Kohli, and R.~Fergus, ``Indoor segmentation and support inference from {RGBD} images,'' in \emph{Eur. Conf. Comput. Vis.}, vol. 7576, 2012, pp. 746--760.

\bibitem{luo2020differentiable}
S.~Luo and W.~Hu, ``Differentiable manifold reconstruction for point cloud denoising,'' in \emph{ACM Int. Conf. Multimed.}, 2020, pp. 1330--1338.

\bibitem{MLS}
D.~Levin, ``The approximation power of moving least-squares,'' \emph{Math. Comput.}, vol.~67, no. 224, pp. 1517--1531, 1998.

\bibitem{Estimating03}
N.~J. Mitra, A.~Nguyen, and L.~J. Guibas, ``Estimating surface normals in noisy point cloud data,'' in \emph{Annu. Symp. Comput. Geom.}, 2003, pp. 322--328.

\bibitem{variants}
G.~Guennebaud and M.~Gross, ``Algebraic point set surfaces,'' \emph{ACM Trans. Graph.}, vol.~26, no.~3, pp. 23.1--23.9, 2007.

\bibitem{Nina}
N.~Amenta and M.~W. Bern, ``Surface reconstruction by voronoi filtering,'' \emph{Discrete Comput. Geom.}, vol.~22, no.~4, pp. 481--504, 1999.

\bibitem{DeyG06}
T.~K. Dey and S.~Goswami, ``Provable surface reconstruction from noisy samples,'' \emph{Comput. Geom. Theory Appl.}, vol.~35, no. 1-2, pp. 124 -- 141, 2006.

\bibitem{Pierre}
P.~Alliez, D.~Cohen-Steiner, Y.~Tong, and M.~Desbrun, ``Voronoi-based variational reconstruction of unoriented point sets,'' in \emph{Eurographics Symp. Geom. Process.}, 2007, pp. 39--48.

\bibitem{Quentin}
Q.~Merigot, M.~Ovsjanikov, and L.~J. Guibas, ``Voronoi-based curvature and feature estimation from point clouds,'' \emph{IEEE Trans. Vis. Comput. Graph.}, vol.~17, no.~6, p. 743–756, 2011.

\bibitem{NormalPei2014}
P.~Luo, Z.~Wu, C.~Xia, L.~Feng, and B.~Jia, ``Robust normal estimation of point cloud with sharp features via subspace clustering,'' in \emph{Int. Conf. Graph. Image Process.}, vol. 9069, 2014, pp. 346--351.

\bibitem{FleishmanCS05}
S.~Fleishman, D.~Cohen{-}Or, and C.~T. Silva, ``Robust moving least-squares fitting with sharp features,'' \emph{ACM Trans. Graph.}, vol.~24, no.~3, pp. 544--552, 2005.

\bibitem{Yoon}
M.~Yoon, Y.~Lee, S.~Lee, I.~P. Ivrissimtzis, and H.~Seidel, ``Surface and normal ensembles for surface reconstruction,'' \emph{Comput.-Aided Des.}, vol.~39, no.~5, pp. 408--420, 2007.

\bibitem{Mederos03}
B.~Mederos, L.~Velho, and L.~H. de~Figueiredo, ``Robust smoothing of noisy point clouds,'' in \emph{SIAM Conf. Geom. Des. Comput.}, 2003.

\bibitem{Wang2013}
Y.~Wang, H.-Y. Feng, F.-{\'E}. Delorme, and S.~Engin, ``An adaptive normal estimation method for scanned point clouds with sharp features,'' \emph{Comput.-Aided Des.}, vol.~45, no.~11, pp. 1333 -- 1348, 2013.

\bibitem{Wang2013Consolidation}
J.~Wang, K.~Xu, L.~Liu, J.~Cao, S.~Liu, Z.~Yu, and X.~D. Gu, ``Consolidation of low-quality point clouds from outdoor scenes,'' \emph{Comput. Graph. Forum}, vol.~32, pp. 207--216, 2013.

\bibitem{Boulch2016}
A.~Boulch and R.~Marlet, ``Deep learning for robust normal estimation in unstructured point clouds,'' \emph{Comput. Graph. Forum}, vol.~35, pp. 281--290, 2016.

\bibitem{Roveri2018}
R.~Roveri, A.~C. Oztireli, I.~Pandele, and M.~H. Gross, ``{PointProNets}: Consolidation of point clouds with convolutional neural networks,'' \emph{Comput. Graph. Forum}, vol.~37, no.~2, p. 87–99, 2018.

\bibitem{Zhou2019}
J.~Zhou, H.~Huang, B.~Liu, and X.~Liu, ``Normal estimation for 3d point clouds via local plane constraint and multi-scale selection,'' \emph{Comput. Aided Geom. Des.}, vol. 129, p. 102916, 2020.

\bibitem{HashimotoS19}
T.~Hashimoto and M.~Saito, ``Normal estimation for accurate 3d mesh reconstruction with point cloud model incorporating spatial structure,'' in \emph{{IEEE/CVF} Conf. Comput. Vis. Pattern Recogn. Workshops}, 2019, pp. 54--63.

\bibitem{PistilliFVM20}
F.~Pistilli, G.~Fracastoro, D.~Valsesia, and E.~Magli, ``Point cloud normal estimation with graph-convolutional neural networks,'' in \emph{{IEEE} Int. Conf. Multimed. {\&} Expo Workshops}, 2020, pp. 1--6.

\bibitem{li2022graphfit}
K.~Li, M.~Zhao, H.~Wu, D.-M. Yan, Z.~Shen, F.-Y. Wang, and G.~Xiong, ``{GraphFit}: Learning multi-scale graph-convolutional representation for point cloud normal estimation,'' in \emph{Eur. Conf. Comput. Vis.}, 2022, pp. 651--667.

\bibitem{Rethinking23}
H.~Du, X.~Yan, J.~Wang, D.~Xie, and S.~Pu, ``Rethinking the approximation error in 3d surface fitting for point cloud normal estimation,'' in \emph{{IEEE/CVF} Conf. Comput. Vis. Pattern Recogn.}\hskip 1em plus 0.5em minus 0.4em\relax {IEEE}, 2023, pp. 9486--9495.

\bibitem{CMG24}
Y.~Wu, M.~Zhao, K.~Li, W.~Quan, T.~Yu, J.~Yang, X.~Jia, and D.~Yan, ``Cmg-net: Robust normal estimation for point clouds via chamfer normal distance and multi-scale geometry,'' in \emph{{AAAI} Conf. Artif. Intell.}\hskip 1em plus 0.5em minus 0.4em\relax {AAAI} Press, 2024, pp. 6171--6179.

\bibitem{MSECNet2024}
H.~Xiu, X.~Liu, W.~Wang, K.~Kim, and M.~Matsuoka, ``Msecnet: Accurate and robust normal estimation for 3d point clouds by multi-scale edge conditioning,'' in \emph{{ACM} Int. Conf. Multimed.}, A.~El{-}Saddik, T.~Mei, R.~Cucchiara, M.~Bertini, D.~P.~T. Vallejo, P.~K. Atrey, and M.~S. Hossain, Eds.\hskip 1em plus 0.5em minus 0.4em\relax {ACM}, 2023, pp. 2535--2543.

\bibitem{SHS23}
Q.~Li, H.~Feng, K.~Shi, Y.~Gao, Y.~Fang, Y.~Liu, and Z.~Han, ``Shs-net: Learning signed hyper surfaces for oriented normal estimation of point clouds,'' in \emph{{IEEE/CVF} Conf. Comput. Vis. Pattern Recogn.}\hskip 1em plus 0.5em minus 0.4em\relax {IEEE}, 2023, pp. 13\,591--13\,600.

\bibitem{li2023NeAF}
S.~Li, J.~Zhou, B.~Ma, Y.-S. Liu, and Z.~Han, ``{NeAF}: Learning neural angle fields for point normal estimation,'' in \emph{AAAI Conf. Artif. Intell.}, 2023.

\bibitem{NeuralGF23}
Q.~Li, H.~Feng, K.~Shi, Y.~Gao, Y.~Fang, Y.-S. Liu, and Z.~Han, ``Neuralgf: Unsupervised point normal estimation by learning neural gradient function,'' in \emph{Adv. in Neural Inf. Process. Syst.}, 2023.

\bibitem{lee2013pseudo}
D.-H. Lee, ``{Pseudo-Label}: The simple and efficient semi-supervised learning method for deep neural networks,'' in \emph{Workshop on Challenges Represent. Learn.}, vol.~3, no.~2, 2013, p. 896.

\bibitem{berthelot2019mixmatch}
D.~Berthelot, N.~Carlini, I.~Goodfellow, N.~Papernot, A.~Oliver, and C.~A. Raffel, ``{MixMatch}: A holistic approach to semi-supervised learning,'' in \emph{Adv. in Neural Inf. Process. Syst.}, vol.~32, 2019, pp. 5049--5059.

\bibitem{lehtinen2018noise2noise}
J.~Lehtinen, J.~Munkberg, J.~Hasselgren, S.~Laine, T.~Karras, M.~Aittala, and T.~Aila, ``{Noise2Noise}: Learning image restoration without clean data,'' in \emph{Int. Conf. Mach. Learn.}, 2018, pp. 4620--4631.

\bibitem{krull2019noise2void}
A.~Krull, T.-O. Buchholz, and F.~Jug, ``Noise2{Void}-learning denoising from single noisy images,'' in \emph{{IEEE/CVF} Conf. Comput. Vis. Pattern Recogn.}, 2019, pp. 2129--2137.

\bibitem{batson2019noise2self}
J.~Batson and L.~Royer, ``{Noise2Self}: Blind denoising by self-supervision,'' in \emph{Int. Conf. Mach. Learn.}, 2019, pp. 524--533.

\bibitem{totaldenoise}
P.~H. Casajus, T.~Ritschel, and T.~Ropinski, ``{Total Denoising}: Unsupervised learning of 3d point cloud cleaning,'' in \emph{{IEEE/CVF} Int. Conf. Comput. Vis}, 2019, pp. 52--60.

\bibitem{Semantic3D}
T.~Hackel, N.~Savinov, L.~Ladicky, J.~D. Wegner, K.~Schindler, and M.~Pollefeys, ``Semantic3d.net: A new large-scale point cloud classification benchmark,'' \emph{ArXiv}, vol. abs/1704.03847, 2017.

\bibitem{SceneNN}
B.-S. Hua, Q.-H. Pham, D.~T. Nguyen, M.-K. Tran, L.-F. Yu, and S.-K. Yeung, ``Scenenn: A scene meshes dataset with annotations,'' in \emph{Int. Conf. 3D Vis. (3DV)}, 2016, pp. 92--101.

\bibitem{modelnet-40}
Z.~Wu, S.~Song, A.~Khosla, F.~Yu, L.~Zhang, X.~Tang, and J.~Xiao, ``{3D ShapeNets}: A deep representation for volumetric shapes,'' in \emph{{IEEE} Conf. Comput. Vis. Pattern Recogn.}, 2015, pp. 1912--1920.

\end{thebibliography}

\end{document}